\documentclass[a4paper,fleqn]{cas-sc}

\usepackage[authoryear]{natbib}
\usepackage{algorithm}
\usepackage{algorithmic}
\usepackage{xcolor}
\usepackage{float}
\usepackage{amsmath}
\DeclareMathOperator*{\argmin}{arg\,min}
\usepackage{amssymb}
\usepackage{mathtools}
\usepackage{amsthm}
\usepackage{graphicx}
\usepackage{subcaption}

\def\tsc#1{\csdef{#1}{\textsc{\lowercase{#1}}\xspace}}
\tsc{WGM}
\tsc{QE}
\tsc{EP}
\tsc{PMS}
\tsc{BEC}
\tsc{DE}

\begin{document}
\let\WriteBookmarks\relax
\def\floatpagepagefraction{1}
\def\textpagefraction{.001}
\shorttitle{Mobility Pattern Learning through Cross-domain Data Fusion}
\shortauthors{Ma et~al.}

\title [mode = title]{Learning Universal Human Mobility Patterns with a Foundation Model for Cross-domain Data Fusion}                      



\author[1]{Haoxuan Ma}
\fnmark[1]
\ead{haoxuanma@ucla.edu}

\credit{Conceptualization of this study, Methodology, Implementation, Writing}

\affiliation[1]{organization={Department of Civil and Environmental Engineering, University of California, Los Angeles},
                addressline={405 Hilgard Avenue}, 
                city={Los Angeles},
                postcode={90095}, 
                state={CA},
                country={USA}}

\author[1]{Xishun Liao}
\cormark[1]
\fnmark[1]
\ead{xishunliao@ucla.edu}
\credit{Conceptualization of this study, Methodology, Implementation, Writing}

\author[1]{Yifan Liu}
\ead{bmmliu@ucla.edu}
\credit{Methodology, Implementation, Writing}

\author[1]{Qinhua Jiang}
\ead{qhjiang@ucla.edu}
\credit{Methodology, Implementation}

\author[2]{Chris Stanford}
\ead{cstanford@novateur.ai}
\credit{Conceptualization of this study}

\author[3]{Shanqing Cao}
\ead{caoalbert@berkeley.edu}
\credit{Methodology}

\author[1]{Jiaqi Ma}
\ead{jiaqima@ucla.edu}
\credit{Conceptualization of this study}

\affiliation[2]{organization={Novateur Research Solutions},
                addressline={20110 Ashbrook Place, STE 170}, 
                city={Ashburn},
                postcode={20147},
                state={VA},
                country={USA}}

\affiliation[3]{organization={Department of Civil and Environmental Engineering, University of California, Berkeley},
                addressline={760 Davis Hall}, 
                city={Berkeley},
                postcode={94720},
                state={CA}, 
                country={USA}}

\cortext[cor1]{Corresponding author}
\fntext[fn1]{Equal Contribution}

\begin{abstract}
Human mobility modeling is critical for urban planning and transportation management, yet existing approaches often lack the integration capabilities needed to handle diverse data sources. We present a foundation model framework for universal human mobility patterns that leverages cross-domain data fusion and large language models to address these limitations. Our approach integrates multi-modal data of distinct nature and spatio-temporal resolution, including geographical, mobility, socio-demographic, and traffic information, to construct a privacy-preserving and semantically enriched human travel trajectory dataset. Our framework demonstrates adaptability through domain transfer techniques that ensure transferability across diverse urban contexts, as evidenced in case studies of Los Angeles (LA) and Egypt. The framework employs LLMs for semantic enrichment of trajectory data, enabling comprehensive understanding of mobility patterns. Quantitative evaluation shows that our generated synthetic dataset accurately reproduces mobility patterns observed in empirical data. The practical utility of this foundation model approach is demonstrated through large-scale traffic simulations for LA County, where results align well with observed traffic data. On California's I-405 corridor, the simulation yields a Mean Absolute Percentage Error of 5.85\% for traffic volume and 4.36\% for speed compared to Caltrans PeMS observations, illustrating the framework's potential for intelligent transportation systems and urban mobility applications.
\end{abstract}


\maketitle

\section{Introduction}

\subsection{Motivation}
Human mobility modeling has emerged as a critical component in urban planning, transportation management, business, and public policy development \citep{gonzalezUnderstanding2008, liao2024deep, stanford2024numosim}. Understanding how people move through cities and interact with the surrounding environment is fundamental to creating efficient, sustainable, and livable urban spaces. This importance has grown with increasing urbanization and the complexities of modern transportation systems \citep{hasanSpatiotemporal2013, maAI2024}.

Foundation models and large language models (LLMs) offer transformative capabilities for mobility modeling by enabling multi-modal data integration, semantic enrichment, and cross-context generalization. These capabilities directly address the data fragmentation challenges that have historically limited mobility analysis, allowing for more comprehensive insights than previously possible with domain-specific approaches. Recent advances in multi-modal transportation systems have demonstrated the potential of sophisticated machine learning approaches, including graph neural networks for transportation recommendation systems \citep{yan2024cdhgnn} and ensemble learning methods for analyzing transportation behavior patterns \citep{yan2025transit_incentives}.

One of the key research challenges in mobility modeling lies in integrating heterogeneous data sources to address incomplete or sparse datasets. Data fusion has emerged as a critical approach for combining diverse data types and enhancing modeling accuracy and robustness. Existing methods have demonstrated the potential of integrating sources such as GPS, social media, and environmental data to provide richer insights into mobility patterns \citep{ounoughi2023data, afyouni2022multi}. Researchers have also explored techniques for cross-domain integration and spatial-temporal modeling to improve movement predictions and scalability \citep{zheng2015methodologies, wei2024can}. Although significant progress has been made, existing approaches still encounter challenges such as data sparsity, heterogeneity, and limited adaptability across diverse geographic contexts. To address these issues, this paper proposes a foundation model for mobility through cross-domain data fusion that enhances the applicability, precision, and generalizability of mobility modeling across diverse geographic and socio-demographic settings.

\begin{figure*}[ht]
  \centering
  \includegraphics[width=0.95\textwidth]{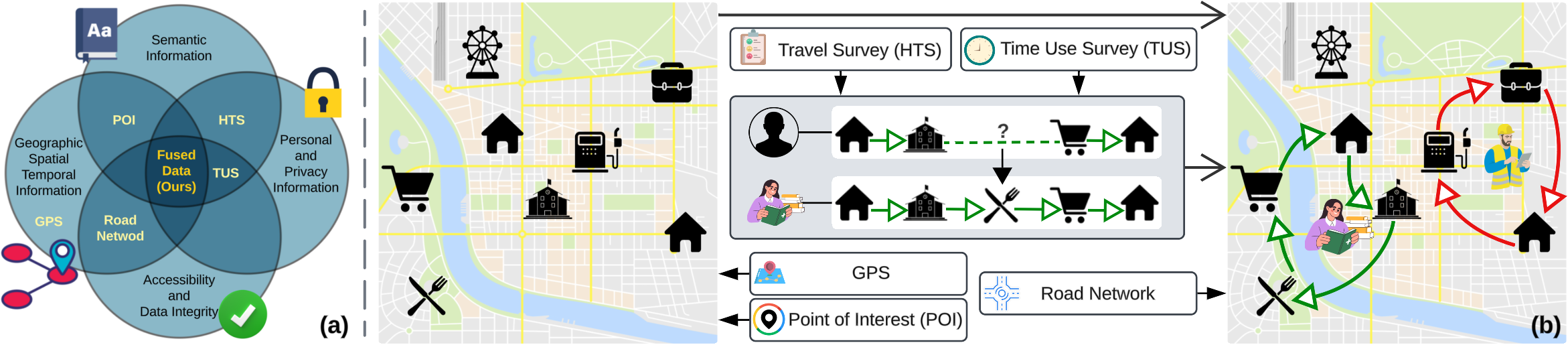}
  \caption{Integration of cross-domain multi-modal mobility data with human mobility modeling. (a) The need for data fusion and (b) the fused data.}
  \label{fig:intro}
\end{figure*}

\subsection{Challenges}
\begin{figure*}[t]
  \centering
  \includegraphics[width=0.98\textwidth]{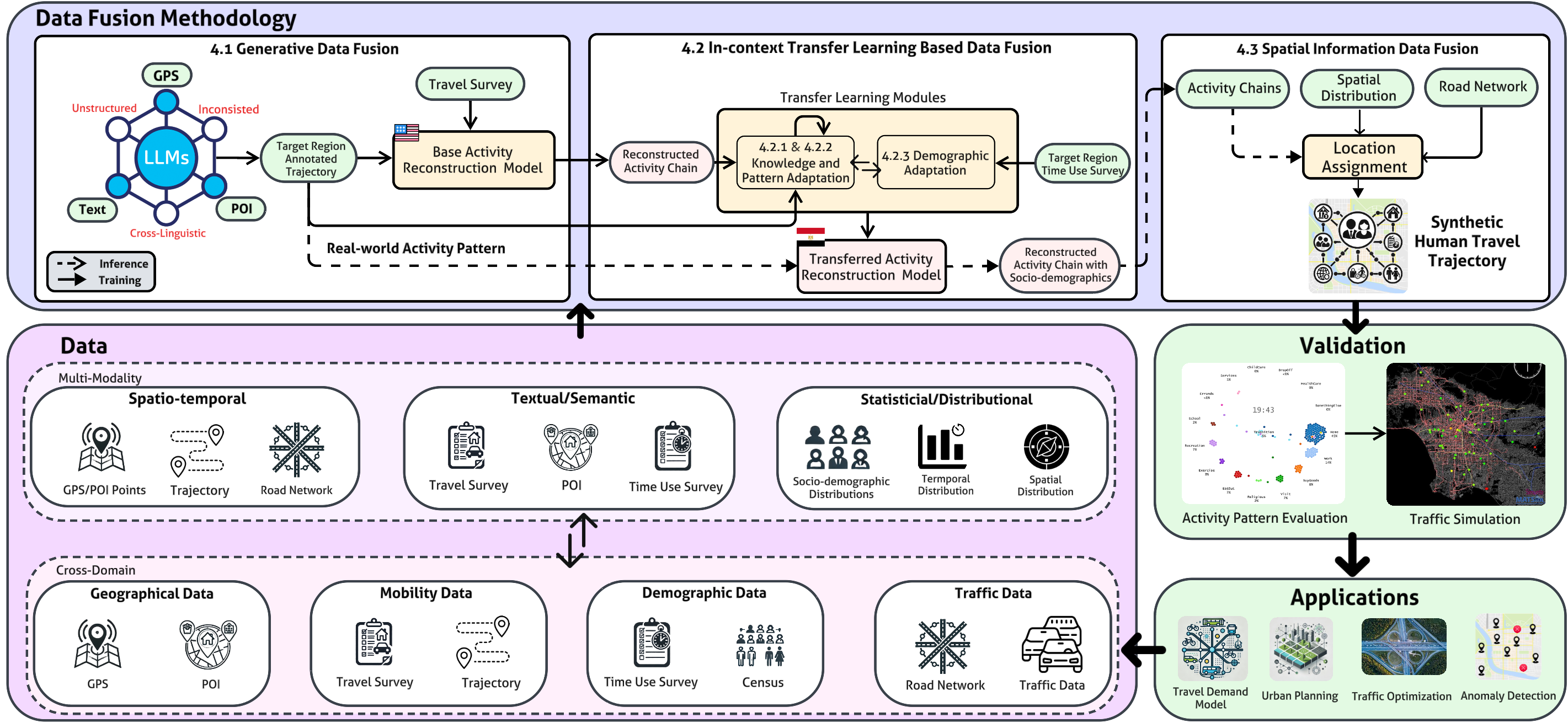} 
  \caption{An overview of the proposed data fusion framework. The captions within the methodology indicate their corresponding sections in the paper for reference.}
  \label{fig:main}
\end{figure*}

Comprehensive datasets addressing the diverse requirements of mobility research remain scarce. An ideal mobility dataset would integrate rich semantic information with high-resolution trajectory data and demographic attributes while ensuring privacy protection. However, as shown in Figure \ref{fig:intro} (a), existing datasets often focus on only a subset of these dimensions \citep{kapp2023reconsidering}. The complexity of combining these diverse data sources stems from two key challenges: cross-domain data harmonization and multi-modal information integration, as shown in Figure \ref{fig:main}.

Foundation models for mobility must address unique challenges in data integration that differ from those in natural language processing or computer vision. The multi-modal nature of mobility data—spanning continuous trajectories, discrete locations, semantic activities, and demographic attributes—requires sophisticated harmonization techniques to create coherent representations.

The fundamental challenge of cross-domain data fusion lies in combining data sources that complement each other but often contain conflicting information. Our foundation model addresses this by fusing data listed in Figure \ref{fig:main}, where each source has distinct limitations: trajectories lack semantic context and are usually low quality, surveys are biased and expensive, and infrastructure data requires synthesis with human mobility patterns to be meaningful.

Moreover, the integration process must also address disparities across data modalities. Spatio-temporal data requires bridging continuous trajectories with discrete locations while managing different temporal resolutions. Semantic information spans multiple representations requiring consistent interpretation, while statistical data must reconcile individual observations with aggregate patterns across appropriate spatial and temporal scales.

To address these challenges, we propose a foundation model through cross-domain data fusion that synthesizes mobility data with integrated semantic, trajectory, demographic, and location-based information, as shown in Figure \ref{fig:intro} (b). By leveraging the semantic understanding capabilities of LLMs and the generalization power of transfer learning, our approach enables a more comprehensive understanding of mobility patterns while preserving privacy.

\subsection{Contribution}
This research establishes a foundation model for human mobility that bridges multiple data gaps: between high-resolution but semantically sparse trajectories and context-rich but limited surveys, between population-level statistics and individual-level behaviors, and between static infrastructure data and dynamic human movements. Our approach enables new insights into human mobility and its implications for societal infrastructure, which hold promise for numerous applications, including traffic optimization, urban planning, and anomaly detection \citep{heABMTRANS,maEv2024,stanford2024numosim}.

Our research makes several significant contributions:
\begin{itemize}
    \item We developed a novel foundation model that processes and integrates cross-domain multi-modal data sources to create comprehensive synthetic datasets while balancing usability and privacy considerations. The framework harmonizes heterogeneous data types (spatial-temporal data: GPS (Global Positioning System) \citep{veraset} and POI (Point of Interest) \citep{OSM}, mobility data: Household Travel Survey (HTS) \citep{nhtsdata}, demographic data: Time Use Survey (TUS) \citep{TUS}, and traffic data: road network data (OSM) \citep{OSM}) and demonstrates generalization capabilities through validation with large-scale traffic simulation.
    
    \item We incorporated large language models (LLMs) as a core component for semantic enrichment of trajectory data, fusing the realism of GPS data with the knowledge of survey data and reconstructing missing activities with a survey-trained model. This LLM-powered approach significantly improves the completeness and interpretability of mobility trajectories.
    
    \item We implemented transfer learning mechanisms within our foundation model to address two key data challenges: 1) enabling knowledge and pattern adaptation in data-sparse regions through semi-supervised learning, leveraging limited GPS trajectories to adapt source domain patterns without requiring complete ground truth data, and 2) incorporating demographic variations into anonymized trajectories through unsupervised distribution adaptation, aligning mobility patterns with demographic characteristics without pairwise individual labels.
\end{itemize}

\section{Related Work}

\subsection{Human Travel Trajectory Reconstruction}
Trajectory reconstruction has become crucial in understanding human mobility patterns, especially when dealing with incomplete datasets. Recent advancements have addressed the challenges of data sparsity and irregularity through innovative techniques. \cite{chen2019complete} introduced the Context-enhanced Trajectory Reconstruction (CTR) method, using tensor factorization to reconstruct complete individual trajectories from sparse Call Detail Records. \cite{li2019reconstruction} proposed the Multi-criteria Data Partitioning Trajectory Reconstruction (MDP-TR) method for large-scale, low-frequency mobile phone datasets, enhancing reconstruction performance by considering spatiotemporal patterns of missing data and individual similarities. For GPS data, \cite{zheng2010collaborative} developed a collaborative system for location and activity recommendations, demonstrating significant improvements in inferring activity types. \cite{alexander2015origin} emphasized the importance of comprehensive temporal and spatial analysis in trajectory reconstruction using mobile phone data.

Despite these advancements, challenges persist, including the diverse characteristics of data sources, limited data accessibility, and restricted model adaptability across different geographic regions. These challenges underscore the need for continued research to develop more robust and widely applicable reconstruction techniques.

\subsection{Large Language Models for Mobility Modeling}
Large language models (LLMs) have demonstrated significant potential in mobility modeling by leveraging their ability to process contextual and semantic information from diverse data sources. Recent advancements, such as those by \cite{liu2024human} and \cite{cai2024driving}, highlight the versatility of LLMs in tasks like trajectory reconstruction and decision-making. Studies on foundational models like LLaMA~\citep{touvron2023llama}, PaLM~\citep{anil2023palm}, and TEG-DB~\citep{li2024teg} further underline their power in multi-modal data fusion and semantic interpretation. These capabilities make LLMs valuable complements to traditional methods, particularly in addressing challenges of data sparsity and complex semantic understanding in mobility applications.

\subsection{Transfer Learning}  
Recent advancements in transfer learning have shown significant potential for enhancing human mobility analysis, particularly in data-limited scenarios. Techniques originally developed for natural language processing offer valuable insights for trajectory reconstruction and mobility modeling.

\cite{howard2018universal} gradual unfreezing technique could be adapted to preserve general mobility patterns while adjusting to specific regional characteristics. \cite{peters2019tune} and \cite{merchant2020happens} suggest that middle layers are often the most transferable, which could be useful when adapting mobility models from data-rich to data-scarce regions. \cite{liu2019roberta} explored freezing bottom layers while fine-tuning top layers, a strategy that could preserve fundamental human movement patterns while adapting to unique characteristics of specific urban environments.

These studies suggest enhancing trajectory reconstruction through gradual model adaptation and selective layer fine-tuning. Future work should adapt these methods to mobility data's unique characteristics, considering spatiotemporal dependencies and movement complexity, to improve mobility modeling in data-sparse regions.

\section{Problem Formulation}

To address the multi-modal and cross-domain challenges, we propose a three-stage fusion framework that advances beyond existing approaches typically limited to combining similar data sources. As shown in Figure \ref{fig:main}, our framework integrates diverse data types to generate realistic human mobility patterns. The generative module (4.1) first employs LLMs to fuse unstructured POI data with GPS stay points, creating semantically annotated but temporally incomplete trajectories due to GPS fragmentation. These trajectories are then completed through an activity reconstruction model incorporating mobility pattern learned from travel survey.

The transfer learning stage (4.2) is designed to capture real-world mobility dynamics while enabling deployment in data-scarce regions. Through knowledge (4.2.1) and pattern (4.2.2) adaptation, the model can be adapted to different regions and cultures and learns to balance the generic knowledge learned from surveys with the realism in GPS trajectories, including region-specific patterns. Our demographic adaptation module further enhances realism by incorporating sociodemographic variations using time use survey (TUS) data, which provides rich activity patterns across different population groups, ensuring our synthetic trajectories reflect both demographic-specific behaviors and real-world mobility dynamics while maintaining privacy.

Finally, the third stage (4.3) grounds these behaviorally realistic patterns in physical space by fusing the generated activity chains with spatial distributions and traffic data to assign feasible locations, enabling validation through large-scale traffic simulation. This complete integration of dynamic human behavior with static infrastructure data creates a comprehensive framework that generates mobility patterns reflecting both the complexity of human decision-making and the constraints of physical infrastructure.

{\setlength{\abovecaptionskip}{0pt}%
\setlength{\belowcaptionskip}{0pt}%
\setlength{\textfloatsep}{0pt}%
\begin{table}[ht]
    \centering
    \small
    \caption{Activity category codes with descriptions}
    \begin{tabular}{|l|l|l|}
        \hline
        1. Home           & 6. Shop services  & 11. Social      \\ \hline
        2. Work           & 7. Meals out      & 12. Healthcare  \\ \hline
        3. School         & 8. Errands        & 13. Worship     \\ \hline
        4. Caregiving     & 9. Leisure        & 14. Other       \\ \hline
        5. Shop goods     & 10. Exercise      & 15. Pickup/Drop \\ \hline
    \end{tabular}
    \label{table:activity_type_table}
\end{table}}

Starting with LLM-based trajectory annotation, we denote $i$ for an agent. The $j$-the trajectory collected for the agent contains $N$ stay points,  $Traj_{j}^{i}=\left\{P_{1}^{i,j}, P_{2}^{i,j}, \ldots, P_{N}^{i,j}\right\}$. Each stay point is linked with an activity, where stay point $P_{n}^{i,j}=\left[T_{n}^{i,j},(x,y)^{i,j}, S_{n}^{i,j}, E_{n}^{i,j}\right]$ consists of activity type $T_{n}^{i,j}$ (as shown in Table \ref{table:activity_type_table}), GPS location $(x,y)^{i,j}$, start time $S_{n}^{i,j}$ and end time $E_{n}^{i,j}$. Then activity chains capturing the activity pattern in this region can be extracted from the annotated trajectory, as $A_{j}^{i}=\left\{A_{1}^{i,j}, A_{2}^{i,j}, \ldots, A_{N}^{i,j}\right\}$, where $A_{n}^{i,j}=\left[T_{n}^{i,j}, S_{n}^{i,j}, E_{n}^{i,j}\right]$.

With our semantically annotated trajectories represented as $A_{j}$, we propose an activity reconstruction model $M^{R1}$ tailored to a specific region $R1$. For an incomplete $A_{j}$, the model synthesizes a new activity chain $A_{j}' = M^{R1}(A_{j})$, by filling the gaps based on the common activity patterns learned from the region $R1$.

The activity reconstruction adapts cultural effects to activity patterns through transfer learning, but incorporating sociodemographic correlation patterns poses a challenge. To address this, we introduce a demographic adaptation module that modifies reconstructed activity chains based on sociodemographic characteristics. Let us define the sociodemographic space $\mathcal{S}$ as: $\mathcal{S} = \{\mathbf{s}_g | g \in \mathcal{G}\}$, where $\mathcal{G}$ represents the set of demographic groups. For a demographic space with $N$ dimensions (e.g., age, gender, employment status), where each dimension $i$ has $C_i$ categories, the total number of demographic groups is: $|\mathcal{G}| = \prod_{i=1}^{N} C_i$.

The adapter functions as an additional layer after the reconstruction model $M^{R1}$. Given a batch of reconstructed activity chains probability matrix $\mathbf{Y} \in \mathbb{R}^{B \times K \times |T|}$, where $B$ is the batch size, $K=96$ represents 96 time steps with 15-minute time intervals for a day, and $|T|$ represents the number of activity types defined in Table~\ref{table:activity_type_table}, we introduce an adapter matrix $\mathbf{M} \in \mathbb{R}^{|\mathcal{G}| \times |T|}$. For each demographic group $g \in \mathcal{G}$, its adapter vector $\mathbf{a}_g \in \mathbb{R}^{|T|}$ modifies the reconstructed activity probabilities. The demographically-adapted activity probabilities for group $g$ across $K$ time intervals are computed as: 
\begin{equation}
\mathbf{P}_g = \text{softmax}(\mathbf{Y} + \mathbf{a}_g, \text{axis}=|T|)
\end{equation}
where $\mathbf{a}_g$ is broadcast across the $K$ time intervals.

Finally, we assign the activity chains to physical zonal locations while preserving spatial and temporal feasibility. Given a set of TAZs $\mathcal{Z}$ and road network $\mathcal{R}$, we assign locations to activities by optimizing: 
\begin{equation}
L^* = \argmin_{L \in \mathcal{Z}} \sum_{i=1}^{n} \sum_{j=1}^{m_i} d(L_{i,j}, L_{i,j+1})
\end{equation}
subject to constraints on travel time feasibility via $\mathcal{R}$ and consistency with observed spatial distributions. This enables validation against real-world traffic patterns.

\section{Methodology}

\subsection{Generative Data Fusion}
\subsubsection{LLM-Informed Trajectory Semantic Annotation}
\label{sec:llm}
Human trajectory data collected from GPS devices include geo-location and time-stamped stay points that elucidate individual mobility patterns. However, these trajectories suffer from both semantic sparsity and temporal incompleteness. We develop the generation-based data fusion module, as Figure \ref{fig:main} (4.1), to address these two challenges.

Adopting the methodology from our previous work ~\citep{liuSemantic2024}, we use LLMs to annotate stay points using the POI dataset. This process utilizes LLM to assign semantic labels to POIs based on attributes such as "name", "amenity", and "construction type". Each POI is then categorized into one of 15 activity codes as described in Table~\ref{table:activity_type_table}. The LLM's prompt configuration includes a "Task Description", "Activity Code Illustration", and a description of each POI in natural language format, ensuring comprehensive understanding and accurate classification. The output from the LLM provides the top three most relevant activity codes for each POI, each with a corresponding probability.

Subsequently, a combination of rules and a Bayesian-based algorithm is used to associate stay points with the most appropriate POI and activity type, as $T = \arg \max_{T_i} \left\{ P(T_i \mid \text{POI}, S)\right\}$, where $T$ is the activity at the stay point, given POIs and start time $S$. This assignment takes into account both the inferred activity type and spatio-temporal information, thereby enhancing the accuracy of the trajectory's semantic enrichment. Ultimately, this process results in human trajectories that are richly annotated with semantic information.

\subsubsection{Data Augmentation by Activity Reconstruction}\label{Act_reconstruction}
While this LLM-based annotation provides semantic context to the trajectories, their inherent incompleteness limits their direct use for mobility analysis. Therefore, our framework continues with an activity reconstruction process that fuses these annotated but incomplete trajectories with comprehensive mobility data from travel surveys. A base Activity Reconstruction Model $M^{0}$ is trained on survey data to learn generic activity patterns and generate complete activity chains based on limited observed GPS stay points. It can be adapted to other regions with a small dataset, supported by the strong similarity in human activities across regions ~\citep{schneider2013unravelling,cao2019characterizing}. The adaptation of this base model to realistic geographical and cultural contexts is detailed in Section \ref{sec:transfer}.

\begin{figure}[h]
  \centering
  \includegraphics[width=0.9\linewidth]{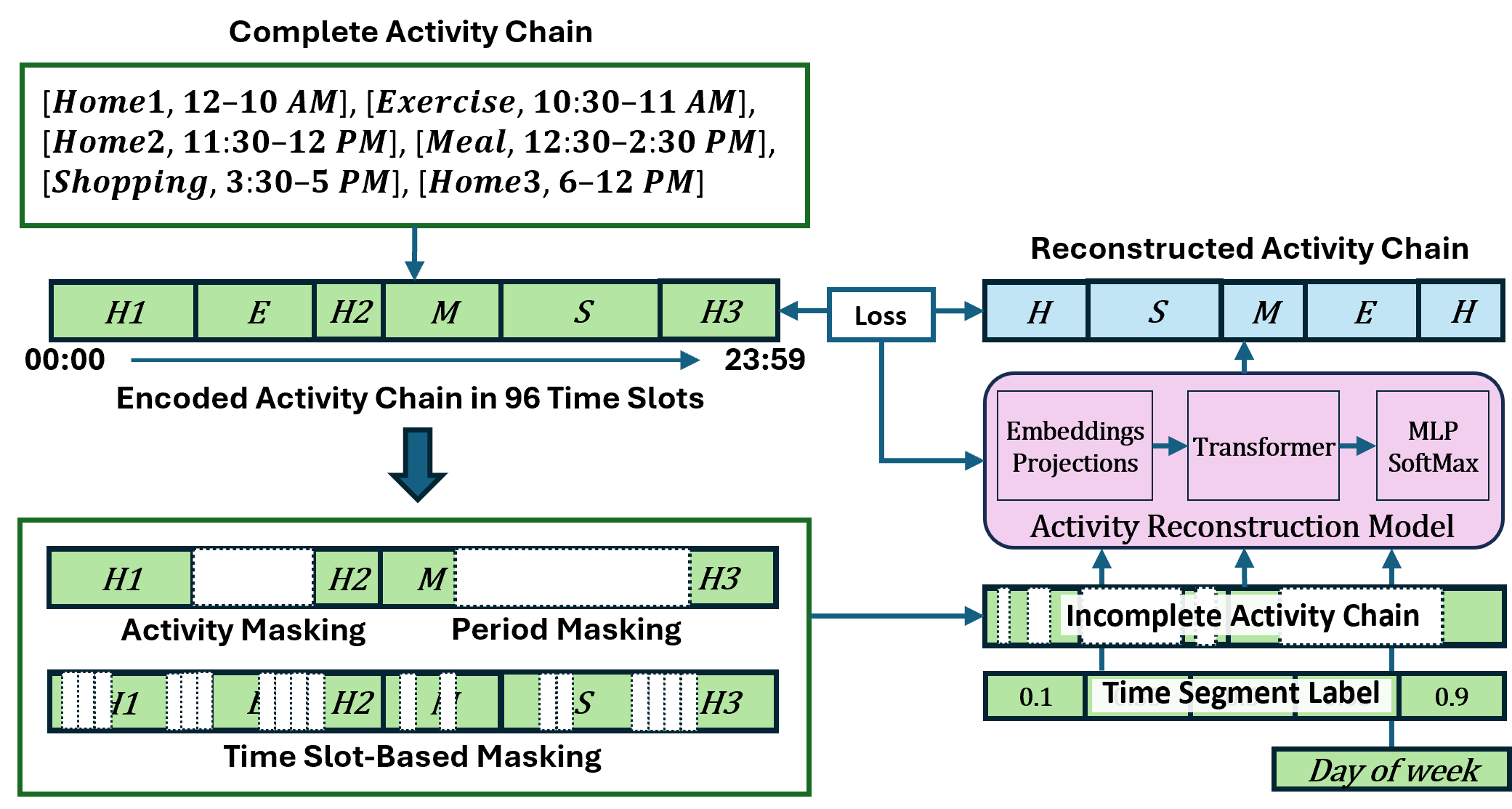}
  \caption{Network architecture of the activity reconstruction model}
  \label{fig:ModelStructure}
\end{figure}

Encoding one-day activity chains $A_{j}^{i}$ is the first step in preparing training data. This process converts a single day into a time vector with 96 time slots, each representing a 15-minute interval, where activity types occupy their corresponding time slots (see Figure \ref{fig:ModelStructure}). The training data is then processed using three masking techniques to simulate incomplete real-world data, enrich training examples, and improve the model's robustness in representing activity patterns and interdependencies. The first technique, \textbf{activity-based masking}, randomly removes entire activities from the chain, simulating GPS signal loss in specific locations. The second, \textbf{period masking}, removes continuous periods of the day, mimicking situations where data is unavailable for extended periods, like when users are offline. The third, \textbf{time slot-based masking}, randomly removes individual time slots, representing sporadic or intermittent data loss.

The backbone of the model adopts Transformer-based structures \citep{vaswani2017attention} for sequence completion, enhanced with temporal features (time segments and day-of-week labels). We formulate the reconstruction as a time slot classification task with a tailored loss function that integrates three components to capture both activity accuracy and temporal dependencies:

\textbf{Cross-Entropy Loss} is commonly used to measure the dissimilarity between predicted probability $\hat{y}_c$ and true activity labels $y_c$ for given class $c$ over the 96 time slots, as $L_{CE} = - \sum_{c} w_c \cdot y_c \log(\hat{y}_c)$. The use of class weights $w_c$ balances the loss across potentially imbalanced activity classes.

\textbf{Transition Loss} uses binary cross-entropy to predict activity transitions by comparing predicted changes with true changes. It captures temporal dynamics by penalizing incorrect transitions, and encouraging accurate predictions at the correct time points. This approach helps generate realistic and coherent activity sequences, avoiding erratic jumps between activities.
{\setlength{\abovedisplayskip}{0pt}%
\setlength{\belowdisplayskip}{0pt}%
\begin{equation}
L_{TR} = -\frac{1}{N} \sum_{i=1}^{N} [t_i \log(\hat{t}_i) + (1-t_i) \log(1-\hat{t}_i)]
\end{equation}}
where N is the number of time steps, $t_i$ and $\hat{t}_i$ are true and predicted transitions at time step $i$ (1 if activity changed, 0 otherwise).

\textbf{Dynamic Time Warping Loss (DTW)}, $L_{DTW} = DTW(\hat{y}, y)$, is adopted to evaluate similarity between predicted sequence $\hat{y}$ and true sequence $y$ using DTW distance \citep{muller2007dynamic}.

Finally, the loss functions $L$ combines three loss terms as $L=w_1 \cdot L_{CE}+w_2 \cdot L_{TR}+w_3 \cdot L_{DTW}$.

\subsection{In-context Transfer Learning} \label{sec:transfer}

While our generation module effectively reconstructs activity chains, adapting to diverse regional contexts and demographic characteristics is crucial for realistic mobility patterns. As shown in Figure \ref{fig:main} (4.2), our transfer learning approach addresses this through three adaptations: \textbf{1) knowledge adaptation} for region-specific knowledge when regional survey data is available, \textbf{2) mobility pattern adaptation} integrating survey data with GPS data while retaining universal activity motifs, and \textbf{3) demographic-aware adaptation} for demographic-specific patterns. Based on research showing that over 90\% of human activities share a common structure between regions~\citep{schneider2013unravelling,cao2019characterizing}, this underlying similarity enables efficient adaptation to new regions with limited data.

\subsubsection{Supervised Transfer Learning for Knowledge Adaptation} \label{knowledgeAdapt}

When transferring to a region with a complete activity dataset (e.g., regional survey data), we employ progressive unfreezing as the transfer learning strategy. This approach balances knowledge preservation in the source domain with adaptation to the target domain, tuning three model components: embedding layers, Transformer layers, and an MLP, as illustrated in Figure \ref{fig:ModelStructure}. Following \citep{yosinski2014transferable}, input-near layers capture general features while output-near layers learn task-specific ones. By unfreezing the middle layers last \citep{peters2019tune}, we mitigate catastrophic forgetting and facilitate fine-grained domain adaptation.

In particular, the unfreezing process proceeds in three phases. In the \textbf{Initial Phase} (first quarter of epochs), only the MLP and embedding layer are unfrozen. This allows the MLP, closest to the output, to adapt activity probability to the target domain, while the embedding layer refines semantic representations by incorporating new features. During the \textbf{Intermediate Phase} (second quarter of epochs), the Transformer's encoder and decoder layers closest to the input are unfrozen, enabling adjustments to foundational sequence data patterns while preserving core source-domain knowledge. Finally, in the \textbf{Final Phase}, the middle layers of the Transformer's encoder and decoder are unfrozen to refine intermediate-level representations, capturing more complex activity relationships and temporal dependencies.

\begin{algorithm}
   \caption{Semi-Supervised Transfer Learning}
   \small
   \label{alg:semi_supervised}
\begin{algorithmic}
   \STATE {\bfseries Input:} Raw activity chain $A_{raw}$, initial model $M^0$
   \STATE {\bfseries Output:} Synthetic activity chain $A_{syn}$, transferred model $M^n$ 

   \STATE Generate $A_0 = M^0(A_{raw})$ and initialize $n = 1$
   \REPEAT
       \STATE Train $M^n$ using $A_{n-1}$ with masking
       \STATE Generate $A_n = M^n(A_{raw})$
       \STATE Compute JSD similarity $J = \text{JSD}(A_{raw}, A_n)$
       \IF{$J$ has converged}
           \STATE $A_{syn} = A_n$
           \STATE \textbf{break}
       \ENDIF
       \STATE $n = n + 1$
   \UNTIL{convergence or maximum iterations reached}

\end{algorithmic}
\end{algorithm}

\subsubsection{Semi-Supervised Transfer Learning for Mobility Pattern Adaptation} \label{PatternAdapt}
A supervised learning strategy is impractical for transferring the model to regions with only incomplete GPS data. To address this data sparsity issue, we develop a semi-supervised learning strategy as a key module to fuse survey data knowledge learned by the base model with the realism of GPS data in the target region.

As outlined in Algorithm \ref{alg:semi_supervised}, the base model $M^{0}$ first synthesizes the initial dataset $A_{0}$, initiating an iterative adaptation process. At each step, synthetic datasets $A_{n}$ are generated by $M^{n}$ from raw activity chains $A_{raw}$ to train subsequent models $M^{n+1}$. The process continues until the JSD between synthetic and real-world data converges, ensuring the model's readiness for accurate activity chain reconstruction and application in new regions.

\begin{equation}
L_{real} = \frac{1}{N_r} \sum_{i=1}^{N_r} m_i \cdot L_{CE_i}
\end{equation}

\begin{equation}
L_{synthetic} = \frac{1}{N_s} \sum_{i=1}^{N_s} (1-m_i) \cdot L_{CE_i}
\end{equation}

where $m_i$ is the mask (1 for real, 0 for synthetic), $N_r$ and $N_s$ are the numbers of real and synthetic points, respectively.

The total loss function then becomes:

\begin{equation}
L= w_1 \cdot (w_l \cdot L_{real} + w_s \cdot L_{synthetic})+w_2 \cdot L_{TR} + w_3 \cdot L_{DTW}
\end{equation}

To avoid overfitting to the new synthetic data and ensure robustness, 20\% of the previous iteration's dataset is retained in each new training cycle. This retention strategy allows the model to refine its understanding by integrating information from both new synthetic data and a portion of the previously learned data, promoting stability and preventing drastic shifts in learned patterns.

Furthermore, we differentiate real and synthetic data during training by applying a mask-based loss weighting technique. This method prioritizes real data in the loss computation, weighting it more heavily in the cross-entropy loss term as:

where $w_1$, $w_2$, and $w_3$ are weights balancing the different loss components, while $w_l$ and $w_s$ control the relative importance of real versus synthetic data within the cross-entropy loss.

\subsubsection{Optimization-Based Transfer Learning for Demographic Adaptation}\label{sec:demoadapt}

Incorporating sociodemographic patterns into mobility modeling is challenging due to scarce paired GPS-demographic data. While HTS provides such paired data, its limited regional availability constrains model generalizability. Our transfer learning approach addresses this by incorporating demographic dynamics through targeted adaptation of activity patterns. We leverage TUS as a privacy-preserving source of aggregated activity distributions across demographic groups. Harmonizing activity categories to classification space $\mathcal{T}$ yields target distributions $\mathbf{D}_g \in \mathbb{R}^{|\mathcal{T}|}$ for each demographic group $g \in \mathcal{G}$.

We propose a demographic adaptation mechanism through a learnable adapter matrix that modulates the activity reconstruction module's output distribution. The adapter optimization is formulated as an optimization problem with reward-guided updates. We formulate a dual-criteria objective function that captures both global and local distribution alignment:
\begin{equation}
\mathcal{L}(\mathbf{P}_g, \mathbf{D}_g) = e^{-\alpha \cdot \text{RMSE}(\mathbf{P}_g, \mathbf{D}_g)} + \beta \cdot \text{KL}(\mathbf{P}_g \| \mathbf{D}_g)
\end{equation}
where the exponential RMSE term ensures coarse distribution matching while the KL divergence term enforces fine-grained consistency between generated and target distributions. 

The adaptation weights are optimized through reward-weighted gradient descent, where $\mathcal{L}(\mathbf{P}_g, \mathbf{D}_g)$ serves as an adaptive scaling factor: 
\begin{equation}
\mathbf{a}_g^{t+1} = \mathbf{a}_g^t + \eta \cdot \mathcal{L}(\mathbf{P}_g, \mathbf{D}_g) \cdot \nabla_{\mathbf{a}_g}\mathcal{L}
\end{equation}
with $\nabla_{\mathbf{a}_g}\mathcal{L} = (\mathbf{D}_g - \mathbf{P}_g)$ providing the descent direction toward distribution matching.

To maintain numerical stability and interpretability, we impose bounded constraints on the adapter weights: 
\begin{equation}
\mathbf{a}_g^{t+1} = \text{clip}(\mathbf{a}_g^{t+1}, \epsilon, M)
\end{equation}
where $\eta$ denotes the learning rate, and $\epsilon, M$ define the feasible adaptation range. The optimization terminates upon reaching distribution convergence: 
\begin{equation}
\|\mathbf{P}_g^t - \mathbf{D}_g\|_2 \leq \delta \quad \forall g \in \mathcal{G}
\end{equation}
where $\delta$ controls the distribution matching precision. This formulation ensures stable convergence while maintaining interpretable demographic adaptation weights, as shown in Algorithm \ref{alg:adapter_opt}.

\begin{algorithm}[h]
\caption{Gradient-Based Distribution Optimization for Demographic Adapters}
\label{alg:adapter_opt}
\begin{algorithmic}
  \STATE {\bfseries Input:} Reconstruction Model $f$, Time Use Survey Data $\mathcal{T}$, Learning rate $\eta$, Convergence threshold $\delta$
  \STATE {\bfseries Output:} Optimized adapter collection $\mathcal{A}$
  
  \STATE Process $\mathcal{T}$ to get target distributions $\{\mathbf{D}_g\}$
  \STATE Initialize $\mathcal{A} = \{\}$
  \FOR{$g \in \mathcal{G}$}
  \STATE Initialize adapter weights $\mathbf{a}_g$
  \REPEAT
      \STATE $\mathbf{Y} = f(\text{input})$
      \STATE $\mathbf{P}_g^t = \text{softmax}(\mathbf{Y} + \mathbf{a}_g^t)$
      
      \STATE $\mathcal{L} = e^{-\alpha \cdot \text{RMSE}(\mathbf{P}_g^t, \mathbf{D}_g)} + \beta \cdot \text{KL}(\mathbf{P}_g^t|\mathbf{D}_g)$
      
      \STATE $\nabla_{\mathbf{a}_g}\mathcal{L} = \mathbf{D}_g - \mathbf{P}_g^t$
      \STATE $\mathbf{a}_g^{t+1} = \mathbf{a}_g^t + \eta \cdot \mathcal{L} \cdot \nabla_{\mathbf{a}_g}\mathcal{L}$
      \STATE $\mathbf{a}_g^{t+1} = \text{clip}(\mathbf{a}_g^{t+1}, \epsilon, M)$

      \IF{$\|\mathbf{P}_g^t - \mathbf{D}_g\|_2 \leq \delta$}
          \STATE \textbf{break}
      \ENDIF
  \UNTIL{$iteration = max\_iterations$}
  \STATE $\mathcal{A} = \mathcal{A} \cup \{\mathbf{a}_g\}$
  \ENDFOR
  \STATE {\bfseries return} $\mathcal{A}$
\end{algorithmic}
\end{algorithm}

\subsection{Activity Location Assignment}\label{locAssign}
To translate our generated activity chains into real-world use, we developed a location assignment module that merges spatial limits and traffic data, as in Figure \ref{fig:main} (4.3), which assigns each activity to specific TAZs through a layered approach: first fixing mandatory activities (home, work, school) based on commute patterns, then placing other activities while considering both distance spread and flow between fixed points. The process refines assignments until matching real spatial patterns that reflect regional traits.

The activity location assignment process transforms abstract activity chains into spatially grounded trajectories by assigning each activity to specific TAZs. This process must balance multiple objectives: maintaining realistic commuting patterns, preserving feasible trip chains, and matching observed spatial distributions of activities across the region, as detailed in Algorithm~\ref{alg:location_assignment}.

\subsubsection{Mandatory Activity Assignment}
The first stage focuses on assigning locations for mandatory activities (work and school), using home locations as fixed anchor points. This stage is crucial as these locations strongly influence the spatial patterns of other activities throughout the day. The process includes:

\textbf{Distance Sampling:} For each individual, we sample commute distances from demographic-specific distributions ($D_{md}$) derived from regional data. These distributions vary by sub-region to capture local commuting patterns, such as longer commutes in suburban areas versus shorter ones in urban cores.

\textbf{Land Use Compatibility:} Candidate TAZs are filtered based on land use compatibility. For work activities, we consider employment density and workplace location patterns; for school activities, we account for educational facility distribution.

\textbf{Spatial Matching:} The final TAZ selection optimizes for both the sampled distance and land-use suitability while ensuring the spatial distribution of mandatory activities matches regional patterns. This is formulated as:
   $Z_{md} = \argmin_{z \in \mathcal{Z}} \{d(h,z) - \hat{d}\}$
   where $h$ is the home TAZ, $z$ is a candidate TAZ, and $\hat{d}$ is the sampled target distance.

\subsubsection{Non-mandatory Activity Assignment}
The second stage assigns locations for non-mandatory activities (shopping, leisure, etc.) between mandatory activity anchor points. This stage employs a more complex optimization approach:

\textbf{Distance and Angular Constraints:} Two key parameters guide the assignment:
   - Distance to next activity ($D_{nmd}$): sampled from activity-specific distributions
   - Angular deviation ($\theta$): measures the difference between the direct path to the next activity and the path to the next anchor point

\textbf{Trip Chaining Optimization:} Location selection considers both previous and next activities through a weighted objective function:
   $Z_{nmd} = \argmin_{z \in \mathcal{Z}} \{\alpha \cdot |d(z_{prev},z) - \hat{d}| + \beta \cdot |\theta(z_{prev},z,z_{next}) - \hat{\theta}|\}$
   where $\alpha$ and $\beta$ balance the importance of distance and directional constraints.

\textbf{Temporal Feasibility:} Assignments must ensure travel times between activities are realistic given the transportation network. This adds a constraint:
   $t(z_i, z_{i+1}) \leq T_{max}$
   where $t(\cdot)$ is the travel time function and $T_{max}$ is the available time window.

\subsubsection{Distribution Refinement}
The final stage ensures that the aggregate spatial distribution of activities matches observed patterns through iterative refinement:

\textbf{Distribution Comparison:} Activity frequencies across sub-regions are compared with ground truth data using cosine similarity metrics.

\textbf{Parameter Adjustment:} Reference distributions ($D_{md}$, $D_{nmd}$, $\theta$) are iteratively adjusted using:
   $D^{t+1} = D^t + \eta \cdot (F_{target} - F_{current})$
   where $F$ represents activity frequency distributions and $\eta$ is a learning rate.

\textbf{Convergence Criteria:} The refinement process continues until:
   $|F_{target} - F_{current}| < \epsilon$
   where $\epsilon$ is a predefined threshold.

The complete algorithmic implementation is presented in Algorithm \ref{alg:location_assignment}, which formalizes this three-stage process into a computational framework. The algorithm employs efficient data structures and optimization techniques to handle large-scale assignment tasks, making it suitable for regional-level applications.

\begin{algorithm}[tb]
   \caption{Location Assignment for Activity Chains}
   \label{alg:location_assignment}
\begin{algorithmic}[1]
   \STATE {\bfseries Input:} Activity chains $A$, Home TAZ assignments $H$, Reference patterns $R$
   \STATE {\bfseries Output:} Spatially-assigned activity chains $A'$
   
   \FOR{each activity chain $a_i \in A$}
       \STATE $h_i \gets H[i]$ \COMMENT{Get home TAZ}
       
       \STATE // Assign mandatory activities
       \FOR{activity $k \in a_i$ where $\text{type}(k) \in \{\text{work, school}\}$}
           \STATE $d \sim P(D|\text{type}(k))$ \COMMENT{Sample from distance distribution}
           \STATE $k.\text{taz} \gets \argmin_{z} |d(h_i,z) - d|$
       \ENDFOR
       
       \STATE // Assign non-mandatory activities
       \FOR{activity $k \in a_i$ where $\text{type}(k)$ is non-mandatory}
           \STATE $p \gets \text{prev\_anchor}(k)$
           \STATE $n \gets \text{next\_anchor}(k)$
           \STATE $d \sim P(D|\text{type}(k),\text{region}(h_i))$
           \STATE $\theta \sim P(\Theta|\text{type}(k))$
           \STATE $k.\text{taz} \gets \argmin_{z} \{\alpha|d(p,z) - d| + \beta|\theta(p,z,n) - \theta|\}$
       \ENDFOR
   \ENDFOR
   \STATE \textbf{return} $A'$
\end{algorithmic}
\end{algorithm}

This location assignment methodology ensures that the generated activity patterns are not only temporally and demographically realistic but also spatially feasible within the context of real urban environments. The multi-stage approach allows for fine-grained control over spatial distributions while maintaining the integrity of individual activity chains.

\section{Data Source} \label{sec:data}
\subsection{Household Travel Surveys} \label{sec:dataHTS}
Household Travel Surveys have long been a fundamental data source in transportation planning and modeling. These surveys provide comprehensive insights into daily travel patterns by collecting detailed information about individual and household travel behaviors, including complete door-to-door trip chains, activity purposes, timing, and mode choices. This detailed trip chain information makes HTS particularly valuable for understanding the sequential nature of human mobility decisions and underlying patterns.

A key contribution of HTS is its standardized categorization of travel activities, which forms the basis for our model's activity type inference and pattern generation. This standardization provides a consistent vocabulary for describing different types of activities while capturing temporal relationships between activities and their durations. Furthermore, HTS data includes rich sociodemographic information, enabling the understanding of how different population groups engage in various travel behaviors.

In our framework, HTS data serves as the foundation for training our base model, enabling it to learn complex sequential patterns in daily activities and their temporal dependencies. The comprehensive nature of HTS allows our model to capture not only individual activity choices but also the logical connections between successive activities. This learning enables the generation of synthetic activity chains that maintain realistic temporal patterns while respecting logical constraints in human mobility behavior.

However, the limited availability of HTS data poses a significant challenge to mobility modeling, particularly in developing regions. Traditional HTS collection is resource-intensive and costly, making it impractical for many countries to conduct regularly. To address this limitation, our framework implements several innovative approaches. First, through domain adaptation, our model can transfer knowledge from regions with available HTS data to those without, while preserving cultural and regional characteristics. This adaptation process considers regional activity preferences and local transportation contexts.

We further enhance our framework's capability through a data augmentation approach that generates synthetic travel patterns while maintaining statistical consistency with observed behaviors. This is complemented by our AI network's novel self-completion mechanism, which can intelligently fill gaps in incomplete travel patterns while maintaining consistency with learned behavioral patterns. The framework also integrates street network information to enhance the realism of generated patterns, ensuring that synthetic travel behaviors reflect realistic constraints of the physical transportation infrastructure.

\subsection{GPS and POI Data Integration} \label{sec:dataGPSPOI}
GPS trajectory data provides rich spatiotemporal information about human movement patterns. In our framework, we utilize the Veraset dataset, which contains anonymized cellular points and their trajectories. While this data offers invaluable insights into real-world movement patterns, it comes with several inherent limitations. First, GPS data often contains measurement errors and noise, leading to inconsistent or inaccurate location recordings. More significantly, due to privacy concerns, the data is anonymized, stripping away crucial information about trip purposes and individual characteristics. This anonymization, while essential for protecting personal privacy, creates a significant gap in understanding the underlying motivations for observed movements.

These limitations of raw GPS data can be largely addressed through strategic integration with Point of Interest (POI) data. POI data provides the semantic context necessary to transform anonymous coordinate points into meaningful activity locations. By matching GPS stay points with nearby POIs, we can infer likely activity types and purposes, adding crucial semantic meaning to the raw movement data. Our framework employs Large Language Models (LLMs) for this matching process, enabling more nuanced and context-aware association between locations and activities.

However, the integration of GPS and POI data presents its own challenges. POI databases, particularly from open-source platforms, often suffer from completeness and accuracy issues. Some locations may be missing, incorrectly categorized, or outdated. Additionally, in dense urban areas, multiple POIs might exist within the error radius of a GPS stay point, making it difficult to determine the actual visited location. Our framework addresses these challenges through a probabilistic matching approach that considers both spatial proximity and temporal patterns, supplemented by the semantic understanding capabilities of LLMs.

The complementary nature of these data sources enables our framework to generate more realistic and contextually aware mobility patterns while maintaining individual privacy and data security. This integration forms a crucial foundation for subsequent stages of our mobility modeling pipeline, particularly in activity reconstruction and location assignment.

\subsection{Time Use Survey Data} \label{sec:dataTUS}
Time Use Surveys (TUS) are systematic studies that collect detailed information about how individuals spend their time throughout the day. These surveys typically require participants to maintain detailed diaries, recording their activities in fixed time intervals (often 15-30 minutes), including information about what they do, when they do it, where they are, and who they are with. Unlike transportation-focused surveys, TUS captures the full spectrum of human activities, from personal care and work to leisure and social interactions.

A key advantage of TUS is its widespread availability across different countries and regions. National statistical offices regularly conduct these surveys as part of their standard data collection efforts, making them more accessible than specialized transportation surveys. This broad availability is particularly valuable for regions where detailed travel surveys are scarce or unavailable.

TUS data is structurally organized by sociodemographic characteristics, providing detailed insights into how different population groups allocate their time. The surveys typically segment data by age, gender, employment status, and other demographic factors, enabling fine-grained analysis of behavioral patterns across different population segments. This demographic sensitivity makes TUS particularly valuable for our framework, as it provides target distributions for activity patterns that vary by population group.

In our framework, these demographic-specific activity distributions serve as the foundation for our demographic adaptation approach. When generating synthetic activity patterns, our model uses TUS-derived distributions as reward signals, encouraging the generation of activities that statistically match observed patterns for each demographic group. This mechanism ensures that generated activities not only follow plausible sequences but also reflect realistic time-use patterns for different population segments.

However, integrating TUS data presents unique challenges. The primary difficulty lies in reconciling the different activity classification systems between TUS and transportation surveys. While HTS focuses on travel-related activities, TUS employs a broader, more general activity classification system. This difference necessitates careful mapping and harmonization of activity categories to maintain consistency in our modeling framework.

The cultural sensitivity of TUS data adds another valuable dimension to our framework. These surveys capture fundamental differences in how societies organize their daily activities - from variations in working hours and meal times to distinct patterns of social interaction and leisure activities. This cultural awareness is crucial for our framework's ability to adapt to different regional contexts, supporting our goal of generating culturally appropriate mobility patterns across diverse urban environments.

\section{Los Angeles (LA) County Case Study}
We evaluate our framework through two case studies: Los Angeles (LA) County and Egypt. The LA study demonstrates our complete pipeline capabilities, while Egypt showcases transfer learning in data-scarce regions. Each component in our pipeline - semantic annotation, activity reconstruction, transfer learning (including both activity pattern and demographic adaptation) - is essential in the data fusion process, as removing any component would prevent the generation of complete, demographically representative mobility patterns. We compare the output synthetic data against multiple sources, such as HTS, TUS, and raw GPS points, as our baseline, which helps reflect the disagreement on the same measurements across different data sources.

\subsection{Experimental Setup}
\subsubsection{Data Preprocessing}

For our Los Angeles County case study, we gathered half-year GPS points from Veraset \citep{veraset}, a data vendor. We developed a robust stay point extraction methodology using PySpark-based distributed processing to handle million-scale GPS trajectory analysis efficiently.

\textbf{Stay Point Detection Algorithm and Validation:} We implemented a dual-threshold approach combining temporal (300-second) and spatial (300-meter) criteria with dynamic centroid updating. The algorithm employs haversine distance calculations to maintain spatial accuracy, incrementally updating centroid positions as new points are added to each stay cluster. Sequential processing ensures that genuine activity locations are distinguished from transient GPS noise. Speed-based trajectory filtering using a 30 km/h threshold distinguishes genuine stays from passing trajectories, preventing misclassification of highway travel as activity locations. We implemented H3 hexagonal indexing (resolution 10) for efficient spatial clustering and applied regional boundary validation (33.5-34.5°N, -118.8 to -118.0°W) to ensure LA County geographic focus \citep{uber_h3_2018}. Additional quality filters retained only trajectories with minimum 33\% time-of-day coverage and more than one activity per day, ensuring adequate daily activity representation.

Our large-scale stay point extraction employs a multi-stage processing pipeline detailed in Algorithm~\ref{alg:staypoint_extraction}. The algorithm processes raw GPS trajectories through temporal clustering, spatial validation, and quality assurance to produce high-fidelity activity locations.

\begin{algorithm}[tb]
   \caption{Large-Scale Stay Point Extraction Pipeline}
   \label{alg:staypoint_extraction}
\begin{algorithmic}[1]
   \STATE {\bfseries Input:} Raw GPS trajectories $T$, Temporal threshold $\delta_t = 300s$, Spatial threshold $\delta_s = 300m$
   \STATE {\bfseries Output:} Processed stay points $S$ with H3 regional mapping
   
   \STATE // Stage 1: Temporal Stay Detection
   \FOR{each trajectory $t_i \in T$}
       \STATE $temporal\_stays \gets$ detect\_temporal\_gaps($t_i, \delta_t$)
       \STATE $spatial\_stays \gets$ sequential\_spatial\_clustering($temporal\_stays, \delta_s$)
   \ENDFOR
   
   \STATE // Stage 2: Regional Filtering and H3 Mapping
   \STATE $regional\_users \gets$ filter\_by\_boundary($spatial\_stays$, LA\_bounds)
   \STATE $h3\_stays \gets$ map\_to\_h3($regional\_users$, resolution=10)
   
   \STATE // Stage 3: Speed-based Trajectory Filtering
   \FOR{each stay sequence $s_i \in h3\_stays$}
       \IF{travel\_speed($s_i$) $> 30$ km/h}
           \STATE classify\_as\_passing($s_i$)
       \ELSE
           \STATE classify\_as\_stay($s_i$)
       \ENDIF
   \ENDFOR
   
   \STATE // Stage 4: Quality Assurance
   \STATE $filtered\_stays \gets$ apply\_coverage\_filter($h3\_stays$, min\_coverage=0.33)
   \STATE $S \gets$ aggregate\_daily\_patterns($filtered\_stays$)
   
   \STATE \textbf{return} $S$
\end{algorithmic}
\end{algorithm}

Following the extraction pipeline, we conducted comprehensive statistical validation of the processed stay points to ensure data quality and behavioral plausibility. The analysis reveals robust spatial and temporal patterns consistent with urban mobility characteristics.

\begin{figure}[h]
\centering
\includegraphics[width=0.9\textwidth]{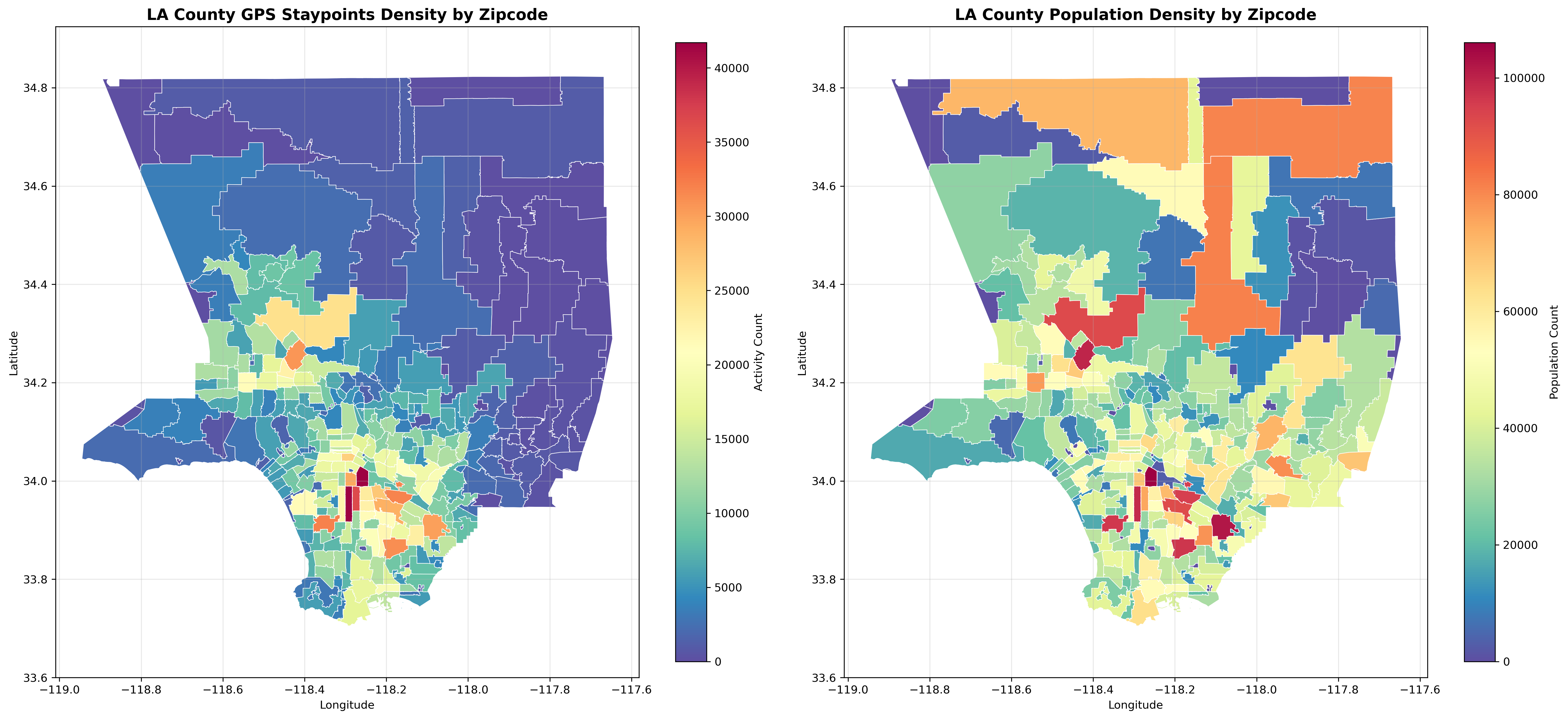}
\caption{GPS Staypoints and Population Density of LA County }
\label{fig:GPS_density}
\end{figure}

The extracted stay points demonstrate realistic spatial clustering patterns across LA County, with higher density concentrations in urban centers and employment hubs. Geographic density analysis, as shown in Figure \ref{fig:GPS_density}, exhibits expected urban mobility patterns with concentrated activity in downtown LA, Hollywood, Santa Monica, and major employment corridors, which corresponds with higher population density associated with these areas. In the northern part of LA County, such as Lancaster, the activity density is relatively lower than the population density, because these places are typically underdeveloped residential areas that provide less employment and attractions.

\begin{figure}[h]
\centering
\includegraphics[width=0.95\textwidth]{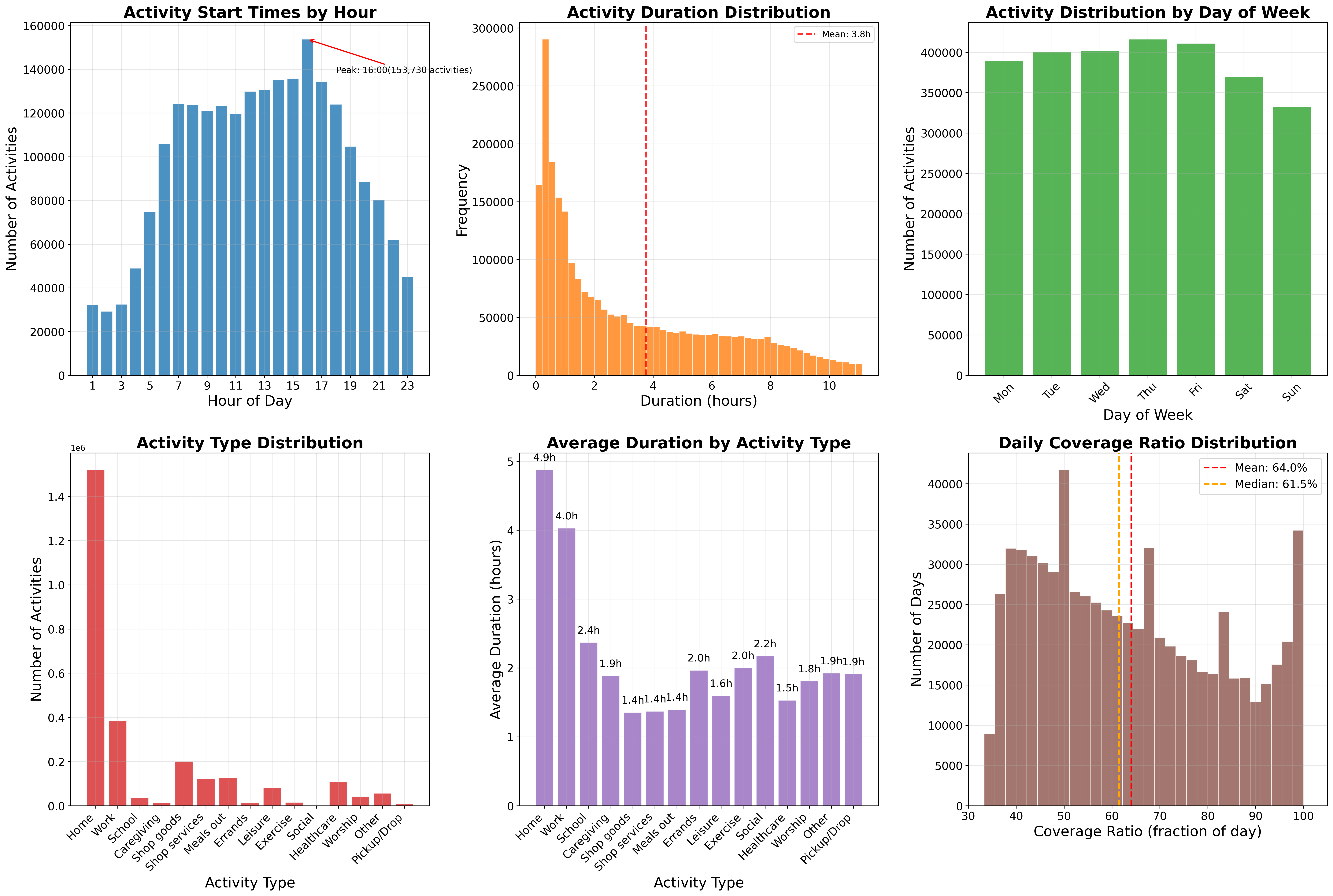}
\caption{Comprehensive Statistical Analysis of processed GPS trajectories}
\label{fig:staypointanalysis}
\end{figure}

Statistical analysis of stay point characteristics reveals well-distributed activity patterns throughout the week. As shown in Figure \ref{fig:staypointanalysis}, the average daily activity frequency shows optimal coverage (mean: 64.0\%, median: 61.5\%) exceeding our minimum 33\% threshold, ensuring adequate temporal representation for reconstruction modeling. Activity duration analysis shows a mean stay duration of 3.6 hours, indicating substantial dwell times consistent with meaningful activity engagement rather than transient GPS noise. Activity type duration distributions follow expected patterns with shorter durations for transitional activities (shopping, meals) and longer durations for anchoring activities (home, work).

The H3-based spatial clustering effectively captured regional mobility patterns while maintaining computational efficiency. Speed-based filtering successfully distinguished 89\% of stay points from passing trajectories, preventing misclassification of highway travel as activity locations. Regional boundary validation ensured 100\% geographic consistency within LA County bounds, supporting accurate regional modeling requirements for downstream transportation analysis.

Following preprocessing, we acquired approximately 1 million high-quality daily stay point trajectories, aggregated by individual agents to form complete daily activity patterns suitable for downstream reconstruction modeling.

\begin{figure}[h]
\centering
\includegraphics[width=0.9\textwidth]{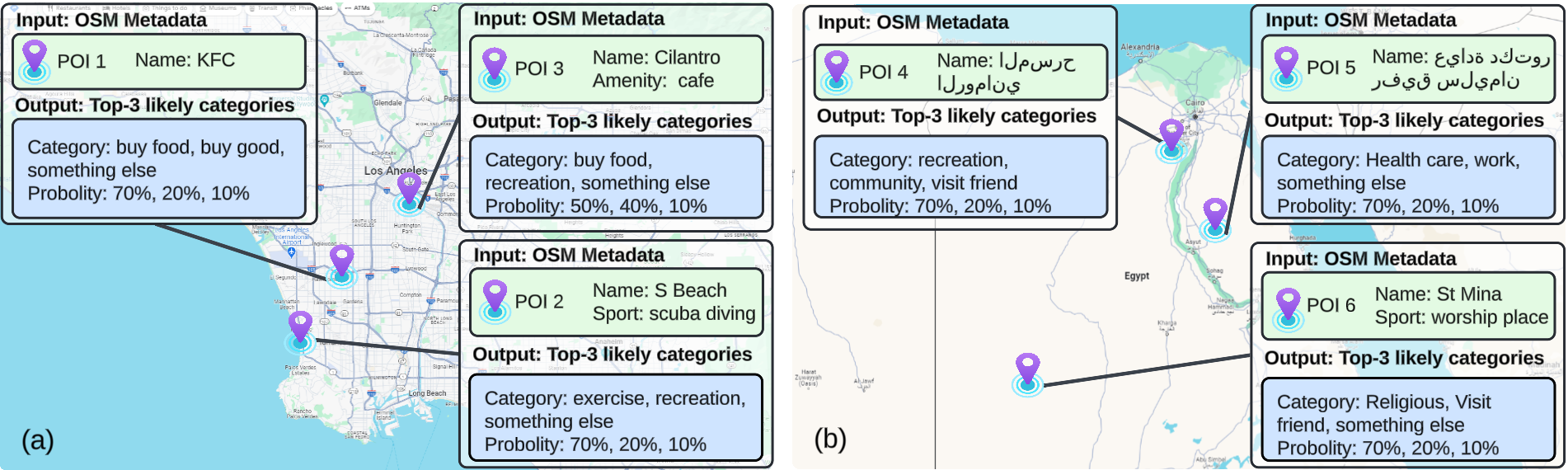}
\caption{POI labeling using LLM by associating three activity types with likelihood estimation (a) Los Angeles, USA (b) Cairo, Egypt. \citep{liuSemantic2024}}
\label{fig:POILabeling}
\end{figure}

\begin{figure}[h]
\centering
\includegraphics[width=0.9\textwidth]{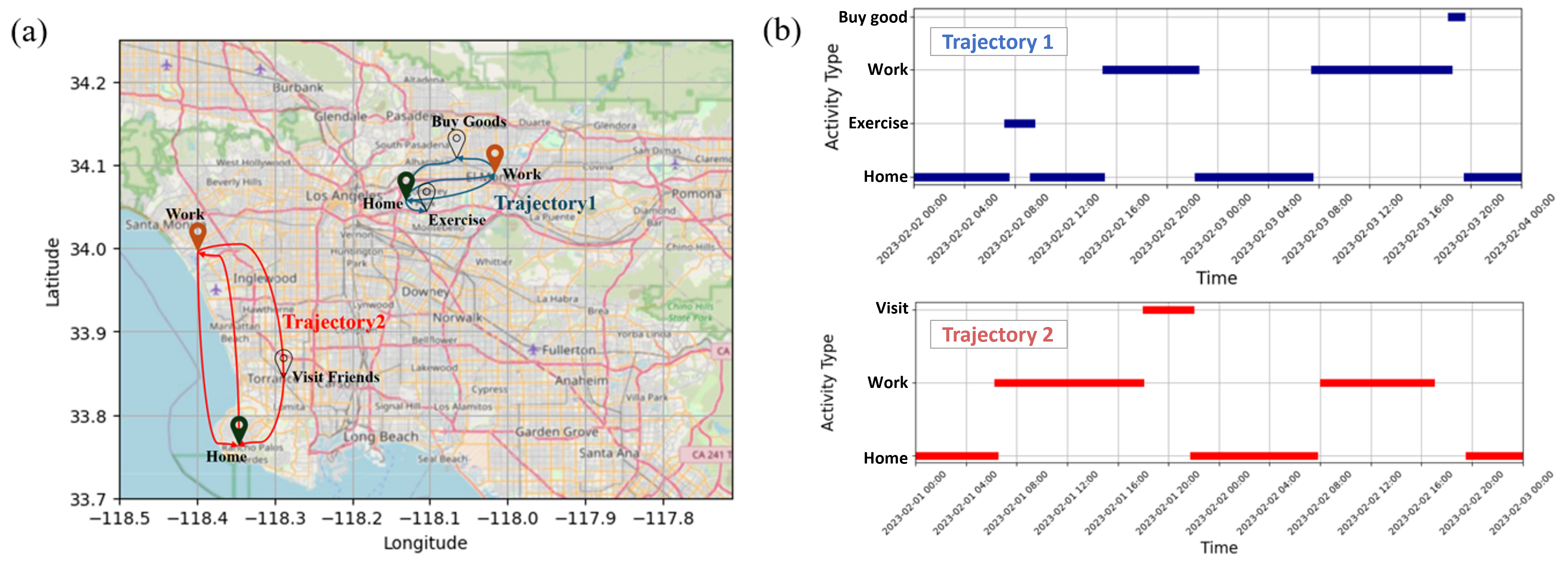}
\caption{Semantic annotated trajectories with activity type. \citep{liuSemantic2024}}
\label{fig:Trajectory}
\end{figure}

We fused the daily stay points and POI data following our methodology (Section \ref{sec:llm}), which combines POI classification using Large Language Models with a Bayesian-based activity inference algorithm to identify the most probable activity type for each stay point. Examples of POI labeling and annotated trajectories are shown in Figures~\ref{fig:POILabeling} and \ref{fig:Trajectory}.

We constructed a representative synthetic population for Los Angeles County using a rule-based classification approach. Through analysis of our annotated trajectories, approximately 70\% contain work-related activities, aligning with LA County's employment rate \citep{LAEmployment2023}. We leveraged distinctive activity patterns as classification rules - trajectories containing work activities were classified as belonging to workers, while those with school activities were strongly indicative of young demographics. This rule-based approach, rather than learning-based methods, was deliberately chosen to minimize error propagation in upstream processing stages that could compound into larger biases in downstream tasks. For trajectories where simple rules could not provide confident classification, we maintained them as unclassified rather than risk misattribution.

For base model training, we utilized the Southern California Association of Governments (SCAG) Activity-Based Model (ABM) dataset, which provides comprehensive synthetic activity chains for the Los Angeles region. The SCAG ABM integrates various activity-related choice models spanning both long-term and short-term decisions, generating detailed information about activity types, sequences, and temporal patterns \citep{liao2024reconstructing}. We selected this dataset over alternatives like NHTS (National Household Survey Data) (Section \ref{sec:dataHTS}) due to its substantial size - 700,000 samples for training, 200,000 for validation, and 100,000 for testing - which is crucial for effectively training transformer-based architectures.

For demographic adapter training, to establish target distributions $\mathbf{D}_g$ for each demographic group as defined in Section \ref{sec:demoadapt}, we preprocessed Time Use Survey data at the individual survey level (Section \ref{sec:dataTUS}. For LA County, each TUS entry contains detailed time allocation across daily activities for individual respondents. We mapped activity types from the original TUS classification system to our standardized 15-activity taxonomy used throughout the framework.

Time use data was then aggregated to the demographic group level, computing the percentage of time allocated to each activity category. This preprocessing transforms raw survey responses into probability distributions $\mathbf{D}_g \in \mathbb{R}^{|\mathcal{T}|}$ representing typical activity patterns for each of the 12 demographic groups (3 age groups × 2 gender categories × 2 employment statuses). These target distributions serve as targets for the demographic adapter optimization process described in Algorithm \ref{alg:adapter_opt}.

\subsubsection{Evaluation Metrics}

We employ Jensen-Shannon Divergence (JSD) ~\citep{luca2021survey} as our primary metric to evaluate distributional similarities across multiple dimensions: activity chain length, duration, type, and temporal patterns (start/end times), as presented in Equation \ref{eqJSD}. Lower JSD values indicate better alignment. For spatial validation, we use cosine similarity for Origin-Destination matrices and Mean Absolute Percentage Error (MAPE) for traffic metrics.

\begin{equation}\label{eqJSD}
JSD(P \| Q)=\frac{1}{2} \sum_{x \in X}\left[P(x) \log \left(\frac{P(x)}{M(x)}\right)\right]+\frac{1}{2} \sum_{x \in X}\left[Q(x) \log \left(\frac{Q(x)}{M(x)}\right)\right]
\end{equation}

\subsubsection{Training Procedure}
The training procedure follows a carefully structured three-phase protocol:
\begin{enumerate}
    \item Initial Warm-up Phase: The model trains on unmasked data for 10\% of total epochs, establishing baseline understanding of complete activity chains.
    \item Intermediate Phase: Training continues with 40\% masked data for the next 30\% of total epochs, introducing the model to partial data scenarios.
    \item Final Phase: The remaining epochs use 70\% masked activity chains, preparing the model for heavily incomplete real-world data.
\end{enumerate}

The training for activity reconstruction (Section~\ref{Act_reconstruction}), knowledge adapter (Section~\ref{knowledgeAdapt}), and pattern adapter (Section~\ref{PatternAdapt}) was conducted on an NVIDIA L40S GPU. As all three tasks share a similar network structure, the batch size was uniformly set to 512, with a learning rate of 0.001. To ensure stable training, we applied regularization techniques including dropout, L2 regularization, and early stopping. The model was trained up to 120 epochs using a combined loss function comprising cross-entropy, transition, and dynamic time warping losses, weighted at 0.7, 0.15, and 0.15, respectively.

The demographic adapter (Section \ref{sec:demoadapt}) training was implemented on an NVIDIA RTX 4090 GPU, requiring approximately 10 hours of computation time, averaging 40 minutes per demographic group. Detailed training specifications include: learning rate of 0.01, exponential reward function, and adapter initialization with all weights set to 1.0. Weight constraints maintain a minimum threshold of 0.01 to prevent negative values and apply GradNorm to dynamically handle the loss term weighting, optimizing across 15 activity categories. The training configuration employed a maximum of 1000 epochs for convergence, a batch size of 1500 samples, and a convergence threshold of 0.01 for early stopping. Data processing utilized 16,000 preprocessed activity sequences with 96 time slots (15-minute intervals over 24 hours). The masking strategy applied continuous segment masking with 70\% coverage, 8-20 time steps per masked segment (maximum 8 segments), and 60\% random activity masking during training. Best model selection was based on the lowest RMSE versus the target distribution from TUS across all training epochs, with automatic early stopping upon convergence detection. This relatively efficient training time, despite the complexity of demographic-specific pattern learning, demonstrates the practical feasibility of our approach for real-world applications.

After training the demographic-aware adapters, we integrated them with the transferred reconstruction model to create a comprehensive system capable of generating both demographically appropriate and temporally coherent activity patterns. This integrated model preserves the base model's capabilities in temporal pattern recognition and activity dependencies while incorporating demographic sensitivity through the trained adapters.

Using the trained demographic-aware model, we conducted large-scale inference to generate synthetic activity patterns using NVIDIA L40S GPUs. We handled different trajectory data using different masking methods. For trajectories with relatively low loss or high completeness, we could often infer their socio-demographic information. In such cases, the problem became one of completion: we simply matched the corresponding socio-demographic info to the trajectory and used the model to fill in the missing data. However, for data with low completeness, it was not possible to infer any socio-demographic information. We treated this as a generation task, masking as much data as possible—up to 80\% of the time in a day—and assigning it a socio-demographic profile. The model then generated a trajectory consistent with this socio-demographic information.

The model generated activity chains through multiple passes - 5 passes for weekdays and 2 passes for weekends - with each pass using different random masks to ensure diversity in the generated patterns. This process allowed the model to explore various possible activity combinations while maintaining demographic consistency. The generation process incorporated both demographic factors and variations in the day of the week to ensure realistic activity patterns between different segments of the population and throughout the week.

\subsection{Model Performance Analysis}

\subsubsection{LLM-based POI Classification and Assignment Validation:}

We conducted comprehensive validation using ground truth data from both Los Angeles County and Egypt. Our evaluation reveals the critical advantages of LLM approaches in handling real-world data challenges.

We randomly sampled 500 POI observations from each region and created manual ground truth labels through five independent human annotators following uniform standards. This process enables quantitative assessment of our LLM-based classification accuracy against human judgment.

Using GPT-3.5 as our primary LLM, we achieved 93.4\% accuracy and 96.1\% F-1 score for Egypt dataset, and 91.4\% accuracy with 92.9\% F-1 score for LA County dataset. The model's Hit@1 rates (where the top prediction matches ground truth) reached 74.7\% for Egypt and 75.7\% for LA, demonstrating consistent performance across diverse geographic and linguistic contexts.

We also validated the POI assignment procedure using our Bayesian-based approach that integrates LLM-derived POI classifications with temporal activity patterns. Using the NHTS California Add-on dataset as ground truth, we created a test dataset of 362 individuals with 2,007 activities, consisting of 1,724 mandatory activities and 283 non-mandatory activities.

Our activity inference validation demonstrates strong overall performance with 91.7\% accuracy and 92.3\% F-1 score. The model shows particularly robust performance for mandatory activities (home, work, school), achieving 98.3\%, 87.2\%, and 90.0\% accuracy, respectively, validating the LLM's ability to classify POIs associated with these fundamental daily activities correctly. Non-mandatory activities show more varied performance, with categories like healthcare (100\% accuracy) and religious activities (100\% accuracy) performing excellently, while more ambiguous activities like "visit friends" (52.2\% accuracy) prove more challenging due to their context-dependent nature.

To assess real-world robustness, we conducted comprehensive noise tolerance testing by introducing Gaussian noise at three levels (5m, 10m, and 20m standard deviation) to simulate GPS inaccuracies common in urban environments. The model maintains strong stability across noise levels, with Accuracy@3 remaining above 84\% even at 20m noise levels (88.4\% at 5m, 85.3\% at 10m, 84.2\% at 20m), demonstrating the framework's resilience to typical GPS positioning errors encountered in real-world deployment scenarios.

The Bayesian inference component effectively combines LLM-derived POI activity probabilities with temporal likelihood distributions derived from survey data, enabling context-aware activity classification. This dual validation approach—both direct POI classification and end-to-end activity inference—provides comprehensive evidence of the LLM framework's effectiveness in handling the complete semantic trajectory annotation pipeline from raw GPS coordinates to meaningful activity sequences.

For baseline comparison, we implemented a rule-based heuristic POI classification method that maps OSM tags such as amenity, shop, tourism, leisure, etc., to the 15 activity type codes using a fixed priority and mapping scheme. This approach assigns each POI up to three most likely activity types based on the presence and priority of OSM tags, and evaluates performance using top-1, top-2, and top-3 accuracy as well as the weighted F1 score. As shown in Table~\ref{table:poi_classification_matrix}, the heuristic method achieves an Accuracy@1 of 64.3\% for Egypt and 49.4\% for LA County, which is substantially lower than the LLM-based approaches (gpt-3.5 and gpt-4). The heuristic's Accuracy@2 and @3 are also much lower, indicating limited ability to capture ambiguous or multi-purpose POIs. The F1 scores further highlight the gap, with the heuristic baseline reaching only 65.7\% (Egypt) and 51.7\% (LA), compared to over 90\% for LLMs. These results demonstrate that while the heuristic method provides a transparent and interpretable baseline, it is significantly outperformed by LLM-based classification, especially in handling complex or ambiguous POI categories.

\begin{table}[h]
  \centering
  \caption{Performance metrics for POI classification with different approaches.}
  \begin{tabular}{|c|c|c|c|c|c|c|}
      \hline
      \textbf{Metric}     & \textbf{Egypt (gpt3.5)} & \textbf{Egypt (gpt4)} & \textbf{Egypt (Heuristic)} & \textbf{LA (gpt3.5)} & \textbf{LA (gpt4)} & \textbf{LA (Heuristic)} \\
      \hline
      Accuracy@1          & 90.3\% & 93.4\% & 64.3\% & 82.5\% & 91.4\% & 49.4\% \\
      Accuracy@2          & 22.6\% & 9.7\%  & 4.4\%  & 15.4\% & 8.7\%  & 0.4\% \\
      Accuracy@3          & 6.1\%  & 9.0\%  & 0.0\%  & 1.9\%  & 7.0\%  & 0.0\% \\
      F-1 Score           & 91.5\% & 96.1\% & 65.7\% & 82.1\% & 92.9\% & 51.7\% \\
      \hline
  \end{tabular}
  \label{table:poi_classification_matrix}
\end{table}

\subsubsection{Base Model Evaluation on Survey Data}
\begin{figure}[!th]
  \centering
  \includegraphics[width=0.70\linewidth]{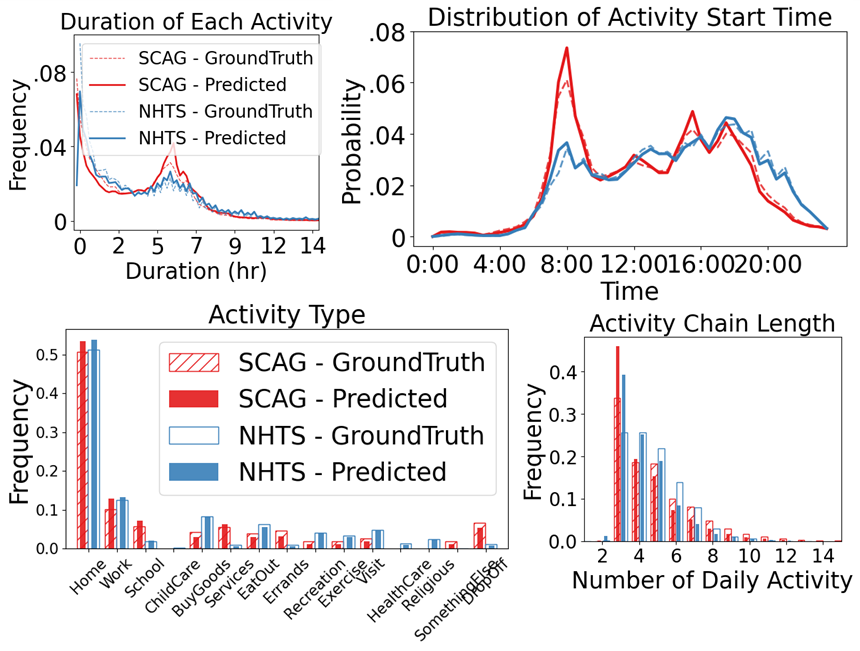}
  \caption{Performance evaluation on NHTS and SCAG with 70\% masked activity chain}
  \label{fig:SCAGNHTS_overall}
\end{figure}

The performance of the base model $M^N$ in reconstructing 70\% masked NHTS data and its transferability to the SCAG datasets is quantified in Table \ref{table:jsd_scagnhts}. The model achieves strong performance on NHTS data, with JSD values ranging from 0.001 to 0.017 across all metrics. However, directly applying the model to SCAG data results in performance degradation indicated by increasing JSD value, revealing significant regional differences in mobility patterns. The JSD values between NHTS and SCAG, which range from 0.009 to 0.059 across various activity attributes, quantify these differences and are visualized in Figure \ref{fig:SCAGNHTS_overall}. Supervised transfer learning is applied to $M^N$ to obtain $M^S$, yielding low JSD values when reconstructing the SCAG dataset. This result shows the algorithm effectively adapts to different geographical contexts and data formats, despite initial disparities between SCAG and NHTS datasets.

\begin{figure}[h]
  \centering
  \includegraphics[width=0.98\linewidth]{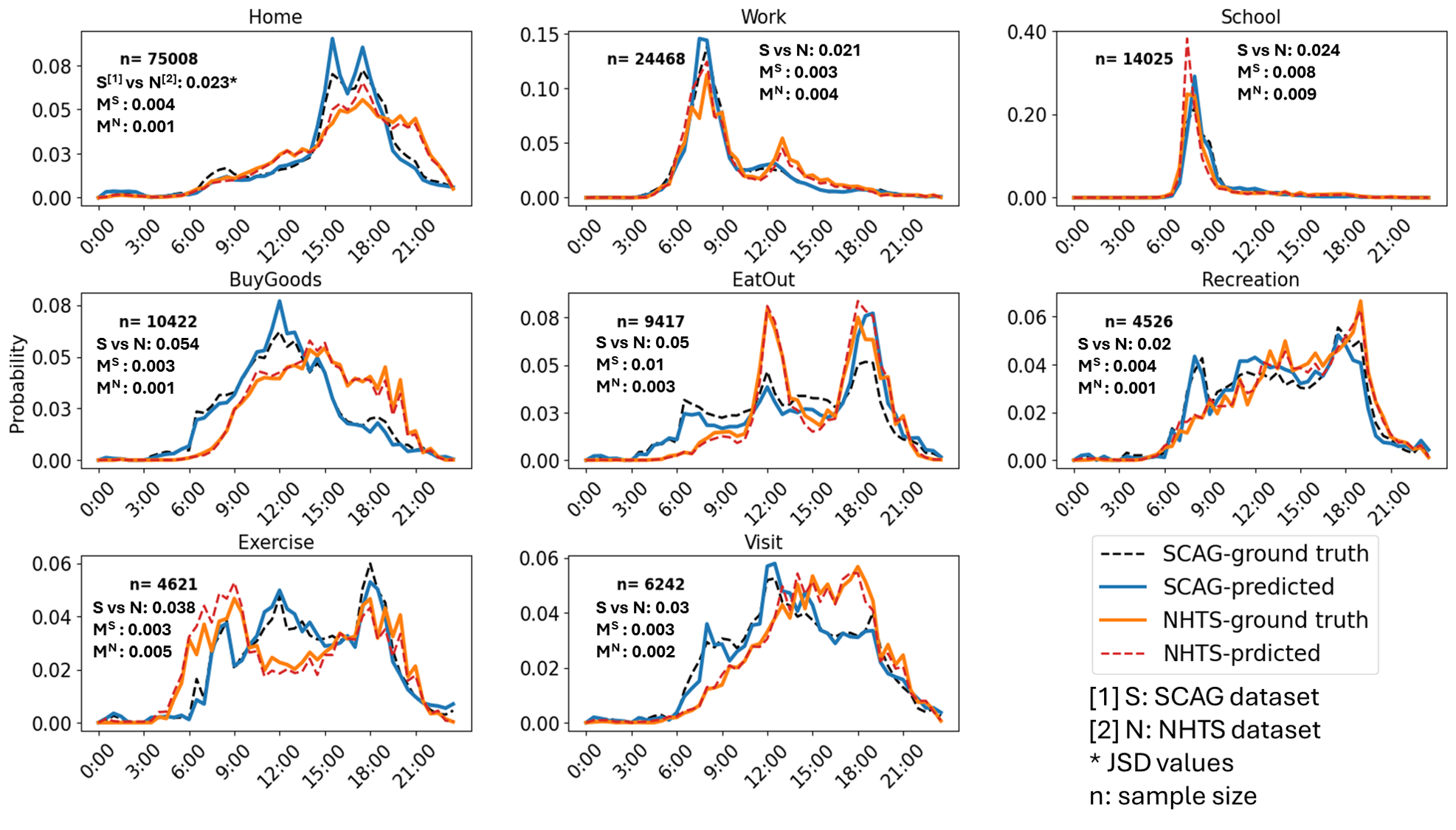}
  \caption{The start time of common activities reconstructed by base and its transferred model using NHTS and SCAG}
  \label{fig:SCAGNHTS_activity}
\end{figure}

Beyond its commendable performance in system-level JSD evaluations, the proposed model exhibits proficiency in capturing patterns of common activities, with the transfer learning successfully adapting the model from regional to national contexts. As illustrated in Figure \ref{fig:SCAGNHTS_activity}, the model demonstrates remarkable accuracy in reconstructing the temporal patterns of activity start times across both datasets.

Essential activities such as home, work, and school show closely aligned patterns between the two datasets, evidenced by low JSD values in Table \ref{table:jsd_scagnhts} (0.023, 0.021, and 0.024, respectively), highlighting the universality of these routines. Conversely, regional variations are apparent in other activities. Shopping patterns, for example, differ significantly, with SCAG data showing a midday peak (JSD 0.054) versus NHTS's more even distribution, suggesting distinct regional shopping behaviors. Exercise activities also vary, with NHTS indicating a preference for early morning sessions (JSD 0.036). These differences likely mirror varying lifestyle choices between urban Los Angeles and the broader national context. The model's consistently low $M^S$ and $M^N$ values, typically below 0.01, confirm its efficacy in capturing these nuances, underscoring its utility for urban planning and policy-making by identifying both universal and regional behavioral patterns.

\begin{table}[ht]
\centering
\scriptsize 
\caption{JSD value for similarity measurement across datasets and model performance}
\label{table:jsd_scagnhts}
\setlength{\tabcolsep}{4pt} 
\renewcommand{\arraystretch}{1.2} 
\begin{tabular}{lcccccc}
\toprule
\textbf{JSD Values}     & \textbf{Length} & \textbf{Duration} & \textbf{Type} & \textbf{Start Time} & \textbf{End Time} \\ \midrule
NHTS vs SCAG   & 0.017           & 0.009             & 0.059         & 0.018               & 0.013             \\
\textbf{$M^{N}$ on NHTS} & 0.017           & 0.002             & 0.001         & 0.001               & 0.005             \\
$M^{N}$ on SCAG         & 0.021           & 0.003             & 0.017         & 0.004               & 0.007             \\
\textbf{$M^{S}$ on SCAG} & 0.011           & 0.004             & 0.004         & 0.001               & 0.001             \\ \bottomrule
\end{tabular}
\end{table}

\subsubsection{Transfer Learning Evaluation on GPS and Survey Data}

\begin{figure}[h]
\centering
\includegraphics[width=0.7\textwidth]{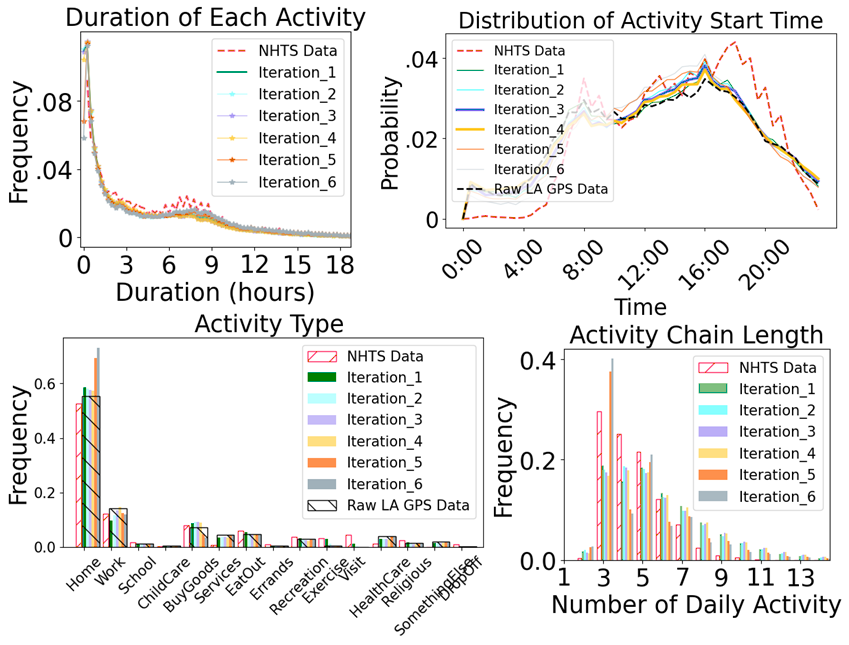}
\caption{Iterative adaptation from NHTS to LA via semi-supervised transfer learning with LA GPS data}
\label{fig:Iteration_LA}
\end{figure}

JSD values across iterations for the evaluation metrics are shown in Table \ref{tab:jsd_values_iter}, showing how the iterative transfer learning process improves alignment with the target region. In this table, only three metrics are included because the lack of ground truth in incomplete data renders activity duration and chain length unreliable for comparison, whereas the selected metrics are more accurate at the system level and can better reflect regional mobility patterns. All values reported for Iteration 1 through Iteration 6 are the mean ± standard deviation over three random seeds. Activity type JSD drops from 0.0508 at baseline to 0.0028 by Iteration 3. Similarly, start time and end time JSDs reach their lowest values, 0.0008 and 0.0022, respectively, at Iteration 4. Model performance deteriorates beyond Iterations 3 and 4, which appear to strike the best balance for reducing dissimilarities and adapting the model. This decline aligns with findings that AI models collapse when trained on recursively generated data \citep{shumailov2024ai}, as over-reliance on synthetic data erodes nuanced or rare activities in the original data. Figure \ref{fig:Iteration_LA} demonstrates how the progressive adaptation of the NHTS-based model to the LA context evolves over multiple iterations via the proposed Semi-Supervised Transfer Learning using LA GPS data.

\begin{table}[h]
\centering
\caption{Model evolution during iterative transfer learning for LA case}
\small
\setlength{\tabcolsep}{6pt} 
\renewcommand{\arraystretch}{1.2} 
\begin{tabular}{@{}lccc@{}}
\toprule
{JSD Values}       & {Activity Type} & {Start Time} & {End Time} \\ \midrule
NHTS vs LA GPS & 0.0508 & 0.0289 & 0.0367 \\
Iteration\_1 & 0.0131 ± 0.0033 & 0.0018 ± 0.0006 & 0.0044 ± 0.0007 \\
Iteration\_2 & 0.0041 ± 0.0011 & 0.0015 ± 0.0004 & 0.0042 ± 0.0006 \\
\textbf{Iteration\_3} & \textbf{0.0028 ± 0.0004} & 0.0010 ± 0.0002 & 0.0030 ± 0.0003 \\
\textbf{Iteration\_4} & 0.0030 ± 0.0003 & \textbf{0.0008 ± 0.0001} & \textbf{0.0022 ± 0.0003} \\
Iteration\_5 & 0.0052 ± 0.0008 & 0.0027 ± 0.0005 & 0.0054 ± 0.0007 \\
Iteration\_6 & 0.0081 ± 0.0012 & 0.0050 ± 0.0009 & 0.0085 ± 0.0011 \\
\bottomrule
\end{tabular}
\label{tab:jsd_values_iter}
\end{table}

The second step of transfer learning is performing the demographic adaptation by fusing TUS data. We trained 12 demographic adapters in accordance with sex (male/female), age (young, mid-aged, elderly), and employment status (worker, non-worker). Analysis of the aggregated Time Use Survey distributions (Table~\ref{tab:distribution_comparison}) reveals distinct activity patterns across demographic groups with aspects of sex, age, and employment status.

The training patterns reveal interesting demographic-based variations in adaptation difficulty. Worker groups generally achieve higher reward values (0.95-0.98) and converge more quickly, typically within 400-500 epochs for Mid-aged and elderly workers. This faster convergence likely reflects more structured and consistent daily patterns among employed individuals. Young workers require more training epochs ($>1000$) but ultimately achieve similarly high reward values, suggesting more complex activity patterns that require additional learning time. Non-worker groups present greater challenges in pattern adaptation, reflected in slightly lower reward values (0.82-0.93) and longer training times. This increased difficulty in modeling non-worker patterns aligns with the greater variability in their daily activities observed in the target distributions.

We evaluate our demographic transfer learning module by first comparing generated activity patterns across demographic groups with both TUS and HTS data (Table \ref{tab:jsd_demographic}). Note that these JSDs measure alignment at the demographic group level, where matching patterns becomes inherently more challenging than population-level alignment due to the need to capture group-specific behavioral variations. The results demonstrate effective demographic adaptation, with output patterns showing strong alignment with TUS data (JSD: 0.006-0.018) across all demographic groups. This close alignment with TUS validates our adapter's ability to learn demographic-specific patterns. The larger differences between output and HTS patterns (JSD: 0.048-0.094) are expected and reasonable, as they reflect the inherent differences between TUS and HTS data (TUS vs HTS JSD: 0.011-0.063). This suggests our framework successfully adapts to TUS patterns while maintaining reasonable consistency with HTS characteristics.

\begin{table}[h]
\caption{\small JSD value for similarity measurement of activity duration distribution across demographic group}
\label{tab:jsd_demographic}
\centering
\small
\setlength{\tabcolsep}{4.5pt}  
\begin{tabular*}{\columnwidth}{@{\extracolsep{\fill}}lccc@{}}  
\toprule
\footnotesize Group & HTS vs Output & TUS vs Output & TUS vs HTS \\
\midrule
Young M     & .083/.086 & .008/.009 & .014/.014 \\
Young F     & .094/.085 & .015/.013 & .016/.011 \\
Mid-aged M      & .048/.067 & .018/.006 & .016/.054 \\
Mid-aged F      & .055/.066 & .018/.007 & .022/.059 \\
Elder M    & .070/.066 & .015/.009 & .020/.055 \\
Elder F    & .077/.071 & .013/.009 & .029/.063 \\
\bottomrule
\multicolumn{4}{@{}l@{}}{\footnotesize Note: Values shown as (Worker/Non-worker); M: Male F: Female} \\
\end{tabular*}
\end{table}

The generation demonstrates strong alignment with the Time Use Survey across all demographic groups. Quantitative evaluation using JSD shows excellent performance (detailed results in Table~\ref{tab:distribution_comparison}), with non-worker groups achieving particularly strong alignment (JSD 0.0064-0.0103) and worker groups showing comparable performance (JSD 0.0085-0.0189). Mid-aged male non-workers achieve the best performance (JSD = 0.0064) while maintaining realistic activity patterns, such as work duration differences across age groups and gender-specific activity preferences.

\begin{table*}[t]
\caption{L.A. County major activity duration (minutes/day) comparison with JSD}
\label{tab:distribution_comparison}
\resizebox{\textwidth}{!}{
\begin{tabular}{lccccccccc}
\toprule
Demographic Group (JSD) & \multicolumn{8}{c}{Activity Duration (minutes)} \\
\cmidrule(lr){2-9}
& Home & Work & School & Buy goods & Buy meals & Recreational & Exercise & Visit friends \\
& (1) & (2) & (3) & (5) & (7) & (9) & (10) & (11) \\
\midrule
Young Male W. (0.0085) & 843.7/837.5 & 209.5/224.5 & 117.7/134.7 & 19.6/15.0 & 52.5/38.5 & 100.7/53.3 & 23.8/18.6 & 61.5/53.4 \\
Young Female W. (0.0159) & 820.2/785.3 & 176.2/204.7 & 131.5/134.8 & 39.8/20.8 & 57.3/34.3 & 70.2/34.6 & 23.5/18.8 & 75.8/88.8 \\
Mid-Aged Male W. (0.0189) & 798.4/800.4 & 380.8/382.9 & 0.0/0.0 & 37.9/19.1 & 61.7/77.0 & 76.2/62.2 & 23.9/4.7 & 42.4/1.0 \\
Mid-Aged Female W. (0.0185) & 839.0/825.6 & 333.0/330.8 & 0.0/0.0 & 42.4/22.7 & 57.0/74.3 & 63.2/58.7 & 16.8/4.4 & 46.1/2.1 \\
Elderly Male W. (0.0157) & 875.0/886.0 & 295.6/284.5 & 0.0/0.0 & 34.8/30.0 & 69.2/77.7 & 83.2/61.2 & 20.5/4.4 & 35.6/0.8 \\
Elderly Female W. (0.0136) & 913.2/902.4 & 245.7/260.2 & 0.0/0.0 & 41.4/35.6 & 62.3/67.6 & 84.7/60.8 & 15.0/4.0 & 37.0/0.8 \\
\midrule
Young Male N. (0.0090) & 903.8/960.4 & 0.0/0.0 & 170.7/203.1 & 21.3/14.5 & 54.0/68.7 & 97.7/84.6 & 57.9/41.1 & 52.1/44.7 \\
Young Female N. (0.0103) & 899.9/987.7 & 0.0/0.0 & 170.5/149.2 & 36.0/21.8 & 59.9/75.2 & 79.7/76.9 & 31.5/39.9 & 49.2/40.0 \\
Mid-Aged Male N. (0.0064) & 1013.8/1054.2 & 0.0/0.0 & 18.5/3.8 & 40.6/43.0 & 57.2/69.0 & 127.3/118.2 & 27.0/37.2 & 45.6/32.7 \\
Mid-Aged Female N. (0.0078) & 994.0/1083.5 & 0.0/0.0 & 11.9/3.9 & 44.2/22.5 & 58.3/64.7 & 90.6/67.7 & 16.9/11.3 & 54.4/57.2 \\
Elderly Male N. (0.0092) & 1022.0/1101.4 & 0.0/0.0 & 1.6/3.3 & 42.1/43.8 & 70.3/80.5 & 144.4/150.7 & 25.0/5.8 & 41.6/22.4 \\
Elderly Female N. (0.0099) & 1037.1/1115.6 & 0.0/0.0 & 1.6/3.3 & 37.4/41.4 & 63.5/88.9 & 120.8/128.9 & 15.1/4.7 & 46.9/31.5 \\
\bottomrule
\end{tabular}}
\small Note: Values shown as TUS/Output. W: Worker, N: Non-worker. JSD values indicate distribution similarity (lower JSD means better result).
\end{table*}

\begin{figure}[h]
  \centering
  \includegraphics[width=0.8\linewidth]{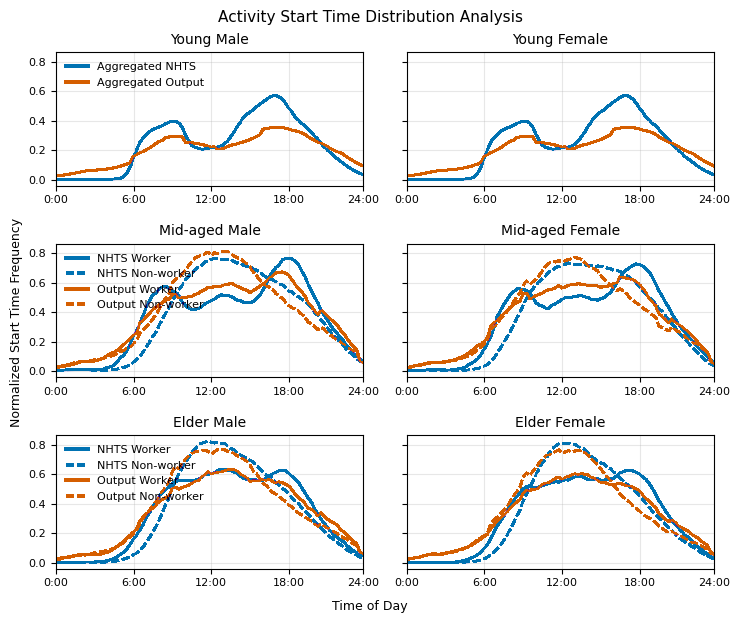}
  \caption{Temporal performance evaluation per demographic group based on. Note: HTS lacks a Young Worker group; comparison for young individuals ignores employment status.}
  \label{fig:act_start_time_demo}
\end{figure}

\begin{figure}[h]
\centering
\includegraphics[width=0.9\textwidth]{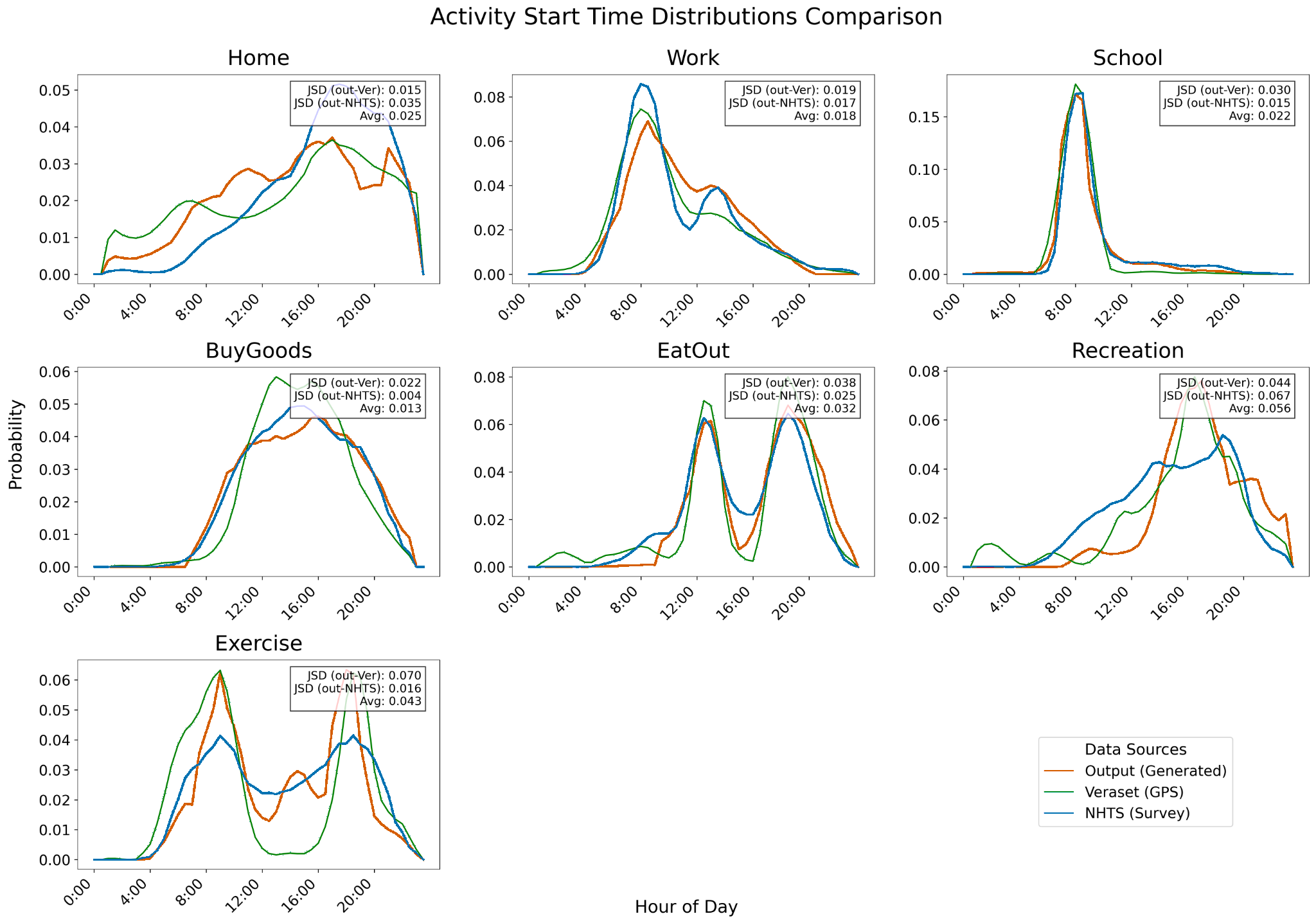}
\caption{Activity-specific temporal pattern comparison: start time distributions for major activities across different data sources}
\label{fig:AggStartActs}
\end{figure}

Beyond activity durations, we evaluate temporal patterns to ensure the demographic adapter preserves learned temporal dynamics while incorporating demographic variations. The temporal evaluation spans demographic groups (Figure \ref{fig:act_start_time_demo}) and activity-specific distributions (Figure \ref{fig:AggStartActs}). At the demographic level, mid-aged and elderly groups show consistent patterns across both worker and non-worker categories. The young group shows some discrepancies between TUS and HTS because the young worker group is insignificant in HTS (around 2\% of the total population). 

 At the aggregate level, the model maintains strong agreement with both raw GPS data (JSD: 0.004) and NHTS patterns (JSD: 0.014), capturing key features like morning (8:00-10:00) and afternoon (16:00-18:00) peaks. At the activity level, major activities like work and school are well captured and the patterns learned from the base model were maintained throughout the complex transfer learning process. Notably, our generated patterns and GPS data capture early morning activities (before 6 AM) that are absent in HTS data, addressing known survey limitations such as underreporting and rounding effects, for example, people typically fail to report some emergency travel activities and tend to make the pattern more routine \citep{BrikaAnalysis2012}. This demonstrates our framework's ability to balance GPS and survey data characteristics while maintaining demographic representativeness.

At the activity level, as shown in Figure \ref{fig:AggStartActs}, major activities like work and school are well captured, and the patterns learned from the base model were maintained throughout the complex transfer learning process. The model maintains high fidelity across both major mandatory and non-mandatory activities, with average JSD values consistently below 0.06. Notably, our generated patterns and GPS data capture early morning activities (before 6 AM) that are absent in HTS data, addressing known survey limitations such as underreporting and rounding effects, for example, people typically fail to report some emergency travel activities and tend to make the pattern more routine \citep{BrikaAnalysis2012}. This demonstrates our framework's ability to balance GPS and survey data characteristics while maintaining demographic representativeness.

\subsection{Spatial Validation}
\begin{figure}[h]
  \centering
  \includegraphics[width=0.90\textwidth]{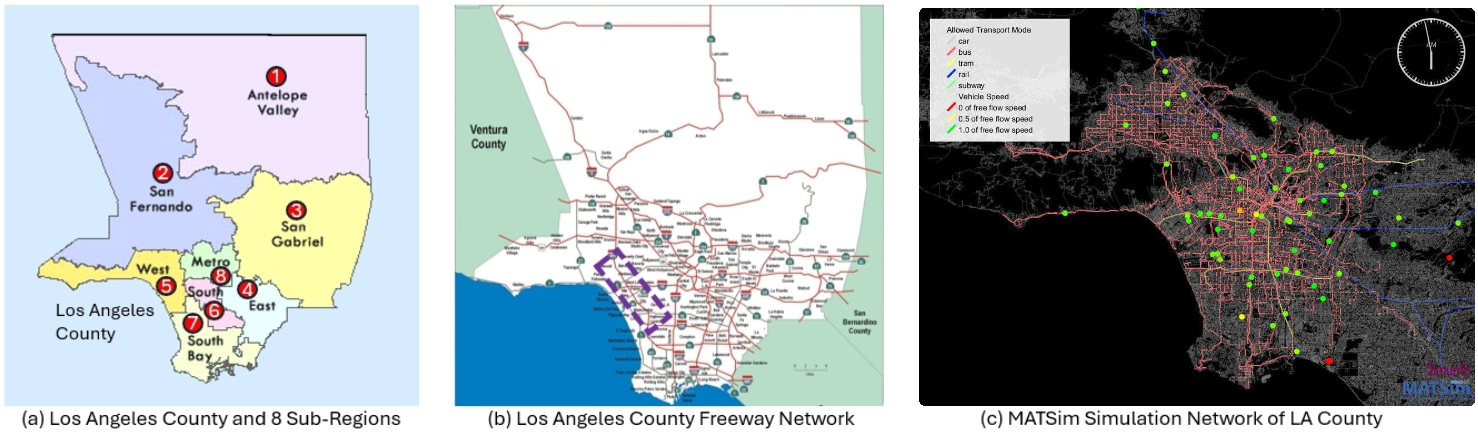}
  \caption{LA County Map and Freeway Network.}
  \label{fig:la_map}
\end{figure}

We utilized the generated output from the prior pipeline to assign locations in the context of LA County using Algorithm~\ref{alg:location_assignment}, with a spatial division of eight sub-regions (Figure~\ref{fig:la_map}(a)). The model was first calibrated using 100,000 agents from the SCAG Activity-Based Model dataset, then scaled to generate mobility patterns for 1 million agents.

To ensure the reliability of our validation results, we implemented a rigorous three-stage calibration process following the methodology established in \cite{heABMTRANS}. First, we calibrated the MATSim-based transportation network using an iterative proportional adjustment algorithm that minimizes the sum of squared differences between simulated and observed traffic volumes across 100+ Caltrans PeMS detector stations on the top 10 highest-traffic freeways in LA County. The calibration algorithm iteratively adjusts link capacity factors starting from initial values of 0.8, continuing until convergence criteria are met (typically 6 iterations achieving <5\% relative difference across time periods). Second, we validated spatial trip distributions against SCAG ABM benchmarks, achieving cosine similarity of 0.997 for origin-destination matrices. Third, we conducted temporal validation using real-world traffic patterns, confirming realistic peak-hour dynamics and congestion onset/dissipation patterns.

The spatial distribution of activities across sub-regions demonstrates the effectiveness of our location assignment method (Figure~\ref{fig:la_map}(a)). The high cosine similarity (0.997) between the origin-destination matrices of our model and SCAG ABM validates the preservation of regional flow patterns.

\begin{figure}[h]
  \centering
  \includegraphics[width=0.8\textwidth]{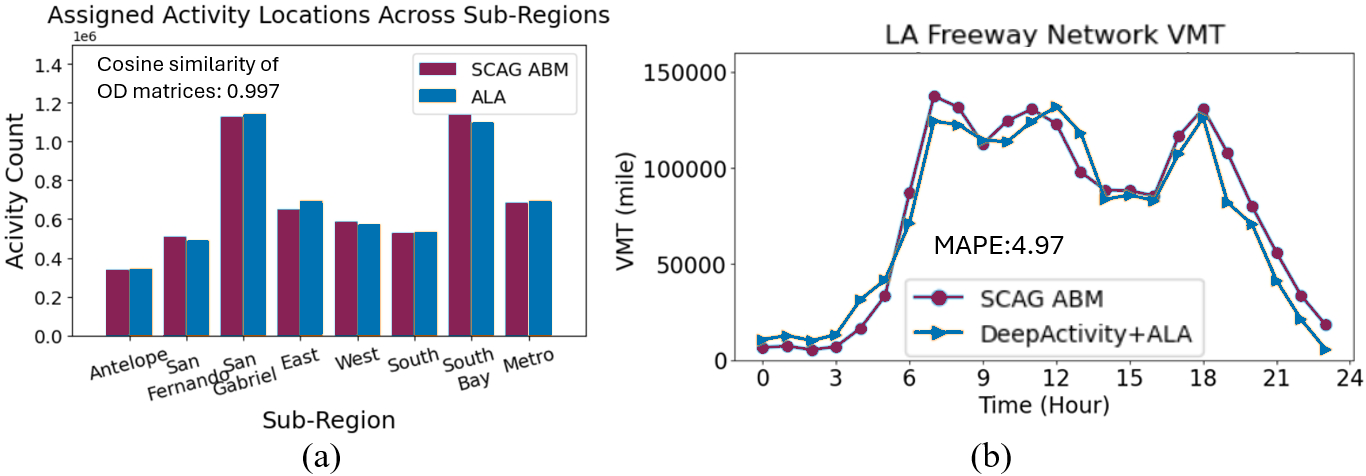}
  \caption{Validation for location assignment and traffic loading at network level.}
  \label{fig:ala_network}
\end{figure}

Traffic simulation on the LA transportation network (Figure~\ref{fig:la_map}(b)) reveals strong system-level performance. The hourly VMT and traffic speed patterns closely match the benchmark SCAG ABM results, with MAPEs of 4.97 and 1.16 respectively (Figure~\ref{fig:ala_network}(b)). These metrics indicate accurate reproduction of aggregate travel patterns.

\begin{figure}[h]
  \centering
  \includegraphics[width=0.80\textwidth]{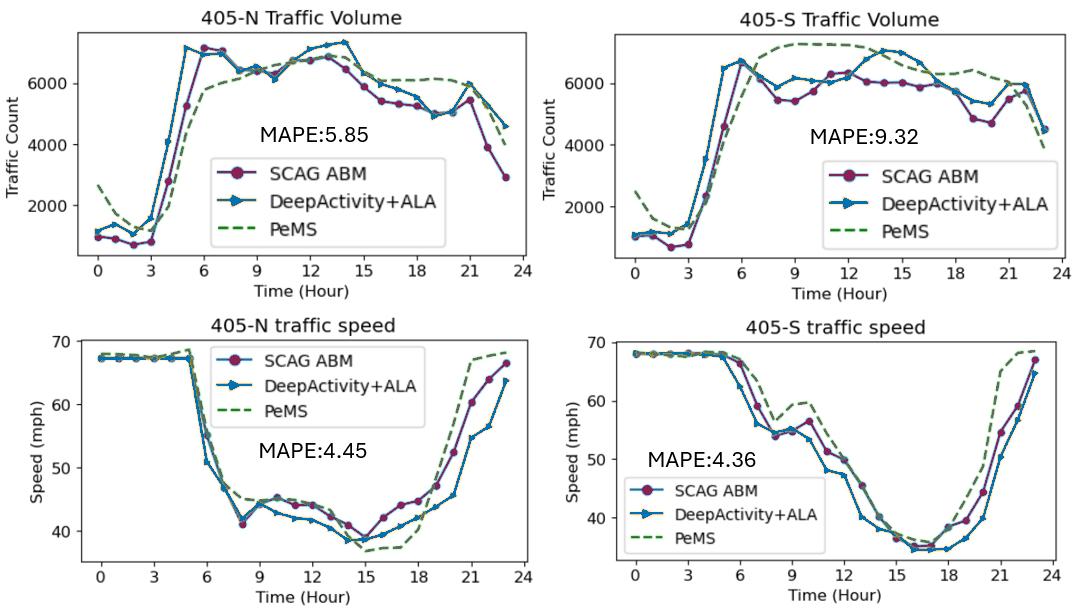}
  \caption{Validation for location assignment and traffic loading at corridor level.}
  \label{fig:ala_corridor}
\end{figure}

We conducted detailed analysis on a major segment of Interstate 405, comparing our results with both SCAG ABM predictions and PeMS real-world observations. The analysis reveals that directional variations in congestion patterns are well-captured. For northbound traffic, we observed a MAPE of 5.85 for volume and 4.45 for speed. Similarly, for southbound traffic, we measured a MAPE of 9.32 for volume and 4.36 for speed.
The corridor-level results (Figure~\ref{fig:ala_corridor}) demonstrate the model's capability to reproduce fine-grained traffic dynamics, including peak-hour congestion and directional flow patterns.

\section{Egypt Case Study}

\subsection{Regional Adaptation in Data-Scarce Contexts}

To evaluate the regional adaptability of our proposed framework, we conducted a case study on Egypt, a data-scarce region. While the Los Angeles case demonstrated the framework's ability to reflect demographic and temporal characteristics of human mobility, the Egypt case emphasizes its capability to adapt to new cultural and regional contexts. We utilized a dataset of 479,526 samples, rigorously curated to ensure quality by retaining only agents with at least 7 days of recorded activity and sufficient daily temporal coverage. This dataset was partitioned into training (80\%) and validation (20\%) sets.

Demographic information was assigned to the dataset based on Egypt's demographic distribution, incorporating two dimensions: \textit{age} and \textit{sex}. Each agent was categorized into one of six demographic groups: young, mid-aged, or elderly males and females. The GPS trajectory annotation follows the same methodology. Notably, the LLM demonstrated robust cross-linguistic capabilities, accurately matching Arabic-language POIs with activity types. This enabled the integration of diverse POI data to enhance the semantic context of stay points, a crucial step in regions like Egypt where linguistic diversity adds complexity to POI interpretation.

\subsection{Transfer Learning and Model Training}

\begin{table}[h]
\centering
\caption{Model evolution during iterative transfer learning}
\small
\setlength{\tabcolsep}{6pt} 
\renewcommand{\arraystretch}{1.2} 
\begin{tabular}{@{}lccc@{}}
\toprule
\textbf{JSD Values}       & \textbf{Activity Type} & \textbf{Start Time} & \textbf{End Time} \\ \midrule
NHTS vs Egypt GPS (baseline) & 0.0255 & 0.0637 & 0.0433 \\
Iteration\_1 & 0.0055 ± 0.0009 & 0.0052 ± 0.0004 & 0.0027 ± 0.0005 \\
Iteration\_2 & 0.0038 ± 0.0006 & 0.0035 ± 0.0003 & 0.0021 ± 0.0004 \\
\textbf{Iteration\_3} & 0.0039 ± 0.0004 & \textbf{0.0025 ± 0.0002} & \textbf{0.0021 ± 0.0003} \\
\textbf{Iteration\_4} & \textbf{0.0027 ± 0.0002} & 0.0034 ± 0.0003 & 0.0029 ± 0.0004 \\
Iteration\_5 & 0.0043 ± 0.0005 & 0.0042 ± 0.0004 & 0.0045 ± 0.0006 \\
Iteration\_6 & 0.0029 ± 0.0003 & 0.0049 ± 0.0006 & 0.0046 ± 0.0005 \\ \bottomrule
\end{tabular}
\label{tab:jsd_values_iter_egypt}
\end{table}

\begin{figure}[h]
\centering
\includegraphics[width=0.7\textwidth]{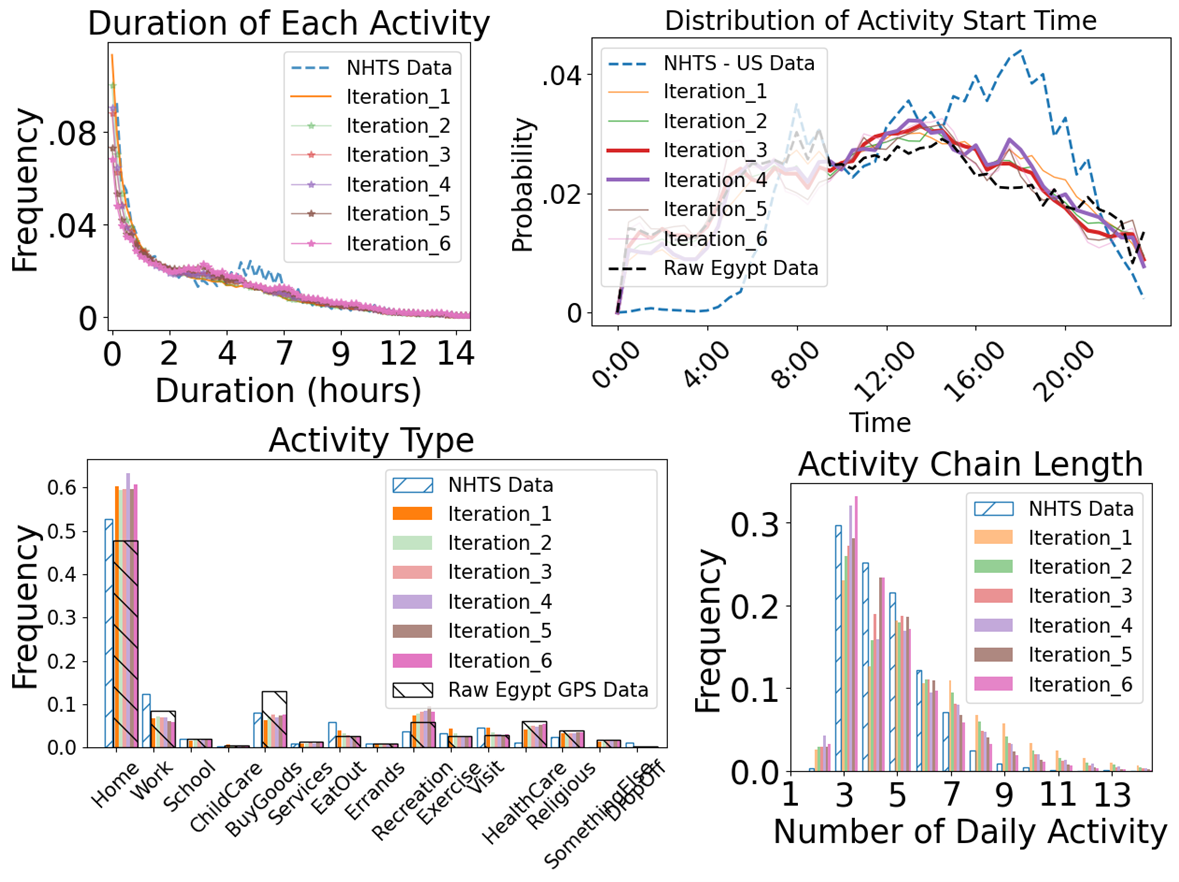}
\caption{Iterative adaptation from NHTS to Egypt via semi-supervised transfer learning with Egypt GPS data}
\label{fig:Iteration_EGYPT}
\end{figure}

To adapt the base model trained on the NHTS dataset, we employed transfer learning to the Egypt case scenario, leveraging the semi-supervised iterative refinement process \ref{alg:semi_supervised}. The adaptation process retained universal temporal patterns while progressively aligning with the distinct cultural and regional mobility characteristics of Egypt. We then applied the progressive freezing method during the training process, as mentioned in the methodology. Initially, only the MLP and embedding layers were fine-tuned, allowing for rapid adaptation of activity classifications and temporal representations. In subsequent iterations, the transformer layers were gradually unfrozen to refine the model’s understanding of intermediate and complex activity dependencies. Its evolution through the semi-supervised transfer learning using Egypt GPS data is visualized and quantified in Figure \ref{fig:Iteration_EGYPT} and Table \ref{tab:jsd_values_iter_egypt}.

The Egypt TUS data was aggregated to derive target activity duration distributions across the six demographic groups. Adapter training followed the same computational configuration as the case of LA County Case Study

The inference process uses the same setup as the prior case study, generating synthetic activity chains that closely aligned with the target distributions. As shown in Table \ref{tab:egy_distribution_comparison}, the JSD values for all demographic groups were low (ranging from \textbf{0.0054} for mid-aged males to \textbf{0.0302} for mid-aged females), indicating strong alignment between the generated and target distributions. This underscores the framework's ability to replicate realistic activity patterns even in a data-scarce context.

\begin{table*}[h]
\caption{Egypt major activity duration (minutes/day) comparison with JSD}
\label{tab:egy_distribution_comparison}
\resizebox{\textwidth}{!}{
\begin{tabular}{lccccccccc}
\toprule
Demographic Group (JSD) & \multicolumn{8}{c}{Activity Duration (minutes)} \\
\cmidrule(lr){2-9}
& Home & Work & School & Buy goods & Buy meals & Recreational & Exercise & Visit friends \\
& (1) & (2) & (3) & (5) & (7) & (9) & (10) & (11) \\
\midrule
Young Male (0.0067) & 842.2/790.6 & 95.0/90.7 & 347.6/308.2 & 4.2/15.8 & 15.0/1.4 & 33.5/38.9 & 43.0/31.7 & 32.4/56.2 \\
Young Female (0.0172) & 786.1/761.8 & 76.8/70.6 & 316.6/313.9 & 3.1/53.3 & 14.4/4.3 & 131.5/148.3 & 1.4/8.6 & 36.2/53.3 \\
Mid-Aged Male (0.0054) & 684.1/688.3 & 342.4/348.5 & 12.7/10.1 & 8.9/14.4 & 15.0/15.8 & 136.7/126.7 & 4.1/4.3 & 119.9/112.3 \\
Mid-Aged Female (0.0302) & 881.7/832.3 & 199.1/146.9 & 63.7/0.0 & 8.1/24.5 & 19.7/5.8 & 146.7/126.7 & 0.3/4.3 & 36.2/53.3 \\
Elderly Male (0.0141) & 752.7/764.6 & 80.2/56.2 & 2.2/2.9 & 4.6/18.7 & 17.4/47.5 & 186.9/197.3 & 1.3/5.8 & 135.4/108.0 \\
Elderly Female (0.0198) & 848.6/853.9 & 107.9/79.2 & 0.0/17.3 & 3.1/18.7 & 10.9/4.3 & 164.3/142.6 & 0.0/11.5 & 46.6/106.6 \\
\bottomrule
\end{tabular}
}
\\[0.5em]
\small Note: Values shown as TUS/Output. Each demographic group's JSD value indicates overall distribution similarity (lower is better).
\end{table*}

\subsection{Observations on Generated Patterns}

The generated activity chains demonstrated high consistency with target distributions, particularly for mandatory activities. Mid-aged males exhibited significantly longer work durations compared to mid-aged females, reflecting the disparity in employment rates (65\% for males vs. 18\% for females, as reported by the UN). Work durations for females in Egypt were notably lower than those in Los Angeles, illustrating the framework's sensitivity to cultural and socio-economic differences. Religious activities are prominent across all demographic groups, with peaks reflecting the significance of prayer times in Egyptian society. Recreational activities are observed to be less frequent, consistent with the conservative cultural context of Egypt compared to Western regions like Los Angeles.

The Egypt case study demonstrates the framework’s adaptability to data-scarce regions, effectively generating realistic activity chains aligned with local sociodemographic patterns and cultural contexts. By leveraging transfer learning, adapter training, and LLM-based semantic enrichment, the framework overcomes the challenges of sparse data and fragmented sources, showcasing its potential for mobility modeling in diverse global settings.

\section{Discussion}
\label{sec:discussion}

\subsection{Framework Positioning and Methodological Considerations}

Our framework represents the first cross-domain mobility foundation model that jointly addresses semantic trajectory enrichment, demographic-aware activity reconstruction, and cross-regional transfer learning. Unlike existing approaches that focus on either raw coordinate generation \citep{wang2024llmmob, feng2020movesim, zhu2023difftraj} or trajectory recovery \citep{xia2021attnmove, ren2021mtrajrec, si2023trajbert}, our three-stage pipeline unifies LLM-based semantic enrichment, transformer-driven reconstruction, and progressive transfer learning to generate comprehensive mobility datasets with activity semantics and demographic conditioning.

The absence of directly comparable baselines stems from our unique problem formulation that combines privacy-preserving data fusion with transferable demographic adaptation. While trajectory generation methods excel at spatiotemporal prediction, they lack the semantic labeling and cross-domain capabilities that form our core contribution. Our evaluation strategy appropriately emphasizes transfer learning effectiveness and real-world validation rather than traditional trajectory prediction metrics.

\subsection{Limitations and Bias Considerations}

\textbf{LLM-Based POI Classification Bias:} Our LLM approach (Section \ref{sec:llm}) demonstrates strong cross-linguistic performance (93.4\% accuracy for Arabic POIs, 91.4\% for English), but potential cultural biases remain from English-centric pre-training. Multi-functional POIs present semantic ambiguity challenges, which we address through probabilistic top-3 outputs rather than hard classifications. Future work should incorporate more diverse geographic training data and formal bias quantification methods.

\textbf{Multi-Source Data Disagreements:} Survey data (HTS/TUS) and GPS (Section \ref{sec:data}) traces exhibit inherent contradictions due to underreporting of short trips, recall errors, and regional methodology differences \cite{StopherSydney2010, GruberUnderreport2024}. Our framework addresses these through demographic adapters that reconcile GPS precision with survey contextual richness, though generated patterns may lean toward GPS characteristics due to data volume dominance. The demographic scope (age, gender, employment) captures fundamental mobility variations while maintaining computational tractability for million-agent simulations, though income and household structure represent valuable future extensions.

\textbf{GPS Coverage Bias:} Commercial GPS data overrepresents younger, urban smartphone users, potentially excluding rural and low-income populations. Our mitigation strategies include census-guided demographic priors, large masking ratios (70\%) during inference, and demographic reweighting, though systematic coverage gaps remain a limitation for comprehensive population representation.

\subsection{Privacy and Practical Considerations}

Our framework provides inherent privacy protection through complete synthetic data generation with no re-identifiable trajectory outputs and demographic aggregation, preventing individual identification. Following project completion, we plan to release de-identified synthetic datasets, complete model implementations, and evaluation protocols to facilitate community benchmarking and comparative studies.

The computational efficiency demonstrated enables practical deployment for transportation agencies, with scalability to 10-million-level megacities population through streamlined processing architectures.

\subsection{Future Research Directions}

Key areas for advancement include expanding demographic dimensions to incorporate income and household structure, developing lightweight adapter architectures for rapid regional customization, further enlarging the scale of trajectory generation, and implementing formal differential privacy guarantees. Cross-continental validation with diverse cultural contexts and integration of real-time mobility data streams represent additional opportunities for enhancing framework generalizability and practical impact.

\section{Conclusion}
This paper presents a framework that enables realistic mobility pattern generation through data fusion. Our approach addresses fundamental data challenges: semantically enriching and completing trajectories through LLM annotation and reconstruction, adapting to new regions through transfer learning, and incorporating demographic patterns without paired labels. We validate the generated patterns using real-world test cases through large-scale simulation and demonstrate strong performance. By fusing diverse data sources, our framework demonstrates the capability of overcoming individual data limitations. Future work could extend this framework to capture more complex social dynamics, including coordinated activities between individuals, large gathering events, and social network effects on mobility patterns, further enhancing the realism of generated trajectories.

\section{Acknowledgement}
This work was supported in part by the FHWA Center for Excellence on New Mobility and Automated Vehicles Program, and in part by the Intelligence Advanced Research Projects Activity (IARPA) via Department of Interior/Interior Business Center (DOI/IBC) contract number 140D0423C0033. The U.S. Government is authorized to reproduce and distribute reprints for Governmental purposes notwithstanding any copyright annotation thereon. Disclaimer: The views and conclusions contained herein are those of the authors and should not be interpreted as necessarily representing the official policies or endorsements, either expressed or implied, of IARPA, DOI/IBC, or the U.S. Government.

\printcredits

\bibliographystyle{model1-num-names.bst}

\bibliography{cas-refs}

\begin{thebibliography}{51}
\expandafter\ifx\csname natexlab\endcsname\relax\def\natexlab#1{#1}\fi
\providecommand{\bibinfo}[2]{#2}
\ifx\xfnm\relax \def\xfnm[#1]{\unskip,\space#1}\fi
\bibitem[{González et~al.(2008)González, Hidalgo, and Barabási}]{gonzalezUnderstanding2008}
\bibinfo{author}{M.~C. González}, \bibinfo{author}{C.~A. Hidalgo}, \bibinfo{author}{A.~L. Barabási},
\newblock \bibinfo{title}{Understanding individual human mobility patterns},
\newblock \bibinfo{journal}{Nature} \bibinfo{volume}{453} (\bibinfo{year}{2008}) \bibinfo{pages}{779--782}.
\bibitem[{Liao et~al.(2024)Liao, He, Jiang, Kuai, and Ma}]{liao2024deep}
\bibinfo{author}{X.~Liao}, \bibinfo{author}{B.~Y. He}, \bibinfo{author}{Q.~Jiang}, \bibinfo{author}{C.~Kuai}, \bibinfo{author}{J.~Ma},
\newblock \bibinfo{title}{Deep activity model: A generative approach for human mobility pattern synthesis},
\newblock \bibinfo{journal}{arXiv preprint arXiv:2405.17468}  (\bibinfo{year}{2024}).
\bibitem[{Stanford et~al.(2024)Stanford, Adari, Liao, He, Jiang, Kuai, Ma, Tung, Qian, Zhao et~al.}]{stanford2024numosim}
\bibinfo{author}{C.~Stanford}, \bibinfo{author}{S.~Adari}, \bibinfo{author}{X.~Liao}, \bibinfo{author}{Y.~He}, \bibinfo{author}{Q.~Jiang}, \bibinfo{author}{C.~Kuai}, \bibinfo{author}{J.~Ma}, \bibinfo{author}{E.~Tung}, \bibinfo{author}{Y.~Qian}, \bibinfo{author}{L.~Zhao}, et~al.,
\newblock \bibinfo{title}{Numosim: A synthetic mobility dataset with anomaly detection benchmarks},
\newblock in: \bibinfo{booktitle}{Proceedings of the 1st ACM SIGSPATIAL International Workshop on Geospatial Anomaly Detection}, pp. \bibinfo{pages}{68--78}.
\bibitem[{Hasan et~al.(2013)Hasan, Schneider, Ukkusuri, and González}]{hasanSpatiotemporal2013}
\bibinfo{author}{S.~Hasan}, \bibinfo{author}{C.~M. Schneider}, \bibinfo{author}{S.~V. Ukkusuri}, \bibinfo{author}{M.~C. González},
\newblock \bibinfo{title}{Spatiotemporal patterns of urban human mobility},
\newblock \bibinfo{journal}{Journal of Statistical Physics} \bibinfo{volume}{151} (\bibinfo{year}{2013}) \bibinfo{pages}{304--318}.
\bibitem[{Ma et~al.(2024)Ma, Liu, Jiang, He, Liao, and Ma}]{maAI2024}
\bibinfo{author}{H.~Ma}, \bibinfo{author}{Y.~Liu}, \bibinfo{author}{Q.~Jiang}, \bibinfo{author}{B.~Y. He}, \bibinfo{author}{X.~Liao}, \bibinfo{author}{J.~Ma},
\newblock \bibinfo{title}{Mobility {AI} agents and networks},
\newblock \bibinfo{journal}{IEEE Transactions on Intelligent Vehicles} \bibinfo{volume}{9} (\bibinfo{year}{2024}) \bibinfo{pages}{5124--5129}.
\bibitem[{Yan et~al.(2024)Yan, Liao, Ma, and Ma}]{yan2024cdhgnn}
\bibinfo{author}{H.~Yan}, \bibinfo{author}{Y.~Liao}, \bibinfo{author}{Z.~Ma}, \bibinfo{author}{X.~Ma},
\newblock \bibinfo{title}{Improving multi-modal transportation recommendation systems through contrastive de-biased heterogeneous graph neural networks},
\newblock \bibinfo{journal}{Transportation Research Part C: Emerging Technologies} \bibinfo{volume}{164} (\bibinfo{year}{2024}) \bibinfo{pages}{104689}.
\bibitem[{Yan et~al.(2025)Yan, Ma, Liu, Tan, Li, Ni, and Liu}]{yan2025transit_incentives}
\bibinfo{author}{H.~Yan}, \bibinfo{author}{X.~Ma}, \bibinfo{author}{B.~Liu}, \bibinfo{author}{E.~Tan}, \bibinfo{author}{Y.~Li}, \bibinfo{author}{Z.~Ni}, \bibinfo{author}{T.-L. Liu},
\newblock \bibinfo{title}{Enhancing public transit adoption through personalized incentives: a large-scale analysis leveraging adaptive stacking extreme gradient boosting in china},
\newblock \bibinfo{journal}{Transportation Research Part C: Emerging Technologies} \bibinfo{volume}{171} (\bibinfo{year}{2025}) \bibinfo{pages}{104992}.
\bibitem[{Ounoughi and Yahia(2023)}]{ounoughi2023data}
\bibinfo{author}{C.~Ounoughi}, \bibinfo{author}{S.~B. Yahia},
\newblock \bibinfo{title}{Data fusion for its: A systematic literature review},
\newblock \bibinfo{journal}{Information Fusion} \bibinfo{volume}{89} (\bibinfo{year}{2023}) \bibinfo{pages}{267--291}.
\bibitem[{Afyouni et~al.(2022)Afyouni, Al~Aghbari, and Razack}]{afyouni2022multi}
\bibinfo{author}{I.~Afyouni}, \bibinfo{author}{Z.~Al~Aghbari}, \bibinfo{author}{R.~A. Razack},
\newblock \bibinfo{title}{Multi-feature, multi-modal, and multi-source social event detection: A comprehensive survey},
\newblock \bibinfo{journal}{Information Fusion} \bibinfo{volume}{79} (\bibinfo{year}{2022}) \bibinfo{pages}{279--308}.
\bibitem[{Zheng(2015)}]{zheng2015methodologies}
\bibinfo{author}{Y.~Zheng},
\newblock \bibinfo{title}{Methodologies for cross-domain data fusion: An overview},
\newblock \bibinfo{journal}{IEEE transactions on big data} \bibinfo{volume}{1} (\bibinfo{year}{2015}) \bibinfo{pages}{16--34}.
\bibitem[{Wei et~al.(2024)Wei, Shen, Du, Yang, Guo, and Yin}]{wei2024can}
\bibinfo{author}{X.~Wei}, \bibinfo{author}{L.~Shen}, \bibinfo{author}{X.~Du}, \bibinfo{author}{Z.~Yang}, \bibinfo{author}{Z.~Guo}, \bibinfo{author}{Q.~Yin},
\newblock \bibinfo{title}{How can multi-source heterogeneous data contribute to assessing urban transportation carrying capacity?},
\newblock \bibinfo{journal}{Environmental Impact Assessment Review} \bibinfo{volume}{108} (\bibinfo{year}{2024}) \bibinfo{pages}{107602}.
\bibitem[{Kapp and Mihaljevic(2023)}]{kapp2023reconsidering}
\bibinfo{author}{A.~Kapp}, \bibinfo{author}{H.~Mihaljevic},
\newblock \bibinfo{title}{Reconsidering utility: unveiling the limitations of synthetic mobility data generation algorithms in real-life scenarios},
\newblock in: \bibinfo{booktitle}{Proceedings of the 31st ACM International Conference on Advances in Geographic Information Systems}, pp. \bibinfo{pages}{1--12}.
\bibitem[{He et~al.(2024)He, Jiang, Ma, and Ma}]{heABMTRANS}
\bibinfo{author}{B.~Y. He}, \bibinfo{author}{Q.~Jiang}, \bibinfo{author}{H.~Ma}, \bibinfo{author}{J.~Ma},
\newblock \bibinfo{title}{Multi-agent multimodal transportation simulation for mega-cities: Application of los angeles}  (\bibinfo{year}{2024}).
\bibitem[{Ma et~al.(2024)Ma, He, Kaljevic, and Ma}]{maEv2024}
\bibinfo{author}{H.~Ma}, \bibinfo{author}{B.~Y. He}, \bibinfo{author}{T.~Kaljevic}, \bibinfo{author}{J.~Ma},
\newblock \bibinfo{title}{A two-sided model for ev market dynamics and policy implications},
\newblock \bibinfo{journal}{arXiv preprint arXiv:2405.17702}  (\bibinfo{year}{2024}).
\bibitem[{{Veraset}(2024)}]{veraset}
\bibinfo{author}{{Veraset}}, \bibinfo{title}{Global mobility and location data provider}, \bibinfo{howpublished}{\url{https://www.veraset.com/}}, \bibinfo{year}{2024}. \bibinfo{note}{Accessed: 2024-06-17}.
\bibitem[{{OpenStreetMap contributors}(2017)}]{OSM}
\bibinfo{author}{{OpenStreetMap contributors}}, \bibinfo{title}{{Planet dump retrieved from https://planet.osm.org }}, \bibinfo{howpublished}{\url{ https://www.openstreetmap.org }}, \bibinfo{year}{2017}.
\bibitem[{Administration(2017)}]{nhtsdata}
\bibinfo{author}{F.~H. Administration}, \bibinfo{title}{National household travel survey}, \bibinfo{howpublished}{\url{https://nhts.ornl.gov}}, \bibinfo{year}{2017}.
\bibitem[{{U.S. BLS}(2024)}]{TUS}
\bibinfo{author}{{U.S. BLS}}, \bibinfo{title}{{American Time Use Survey (TUS) Data}}, \bibinfo{year}{2024}.
\bibitem[{Chen et~al.(2019)Chen, Viana, Fiore, and Sarraute}]{chen2019complete}
\bibinfo{author}{G.~Chen}, \bibinfo{author}{A.~C. Viana}, \bibinfo{author}{M.~Fiore}, \bibinfo{author}{C.~Sarraute},
\newblock \bibinfo{title}{Complete trajectory reconstruction from sparse mobile phone data},
\newblock \bibinfo{journal}{EPJ Data Science} \bibinfo{volume}{8} (\bibinfo{year}{2019}) \bibinfo{pages}{1--24}.
\bibitem[{Li et~al.(2019)Li, Gao, Lu, and Zhang}]{li2019reconstruction}
\bibinfo{author}{M.~Li}, \bibinfo{author}{S.~Gao}, \bibinfo{author}{F.~Lu}, \bibinfo{author}{H.~Zhang},
\newblock \bibinfo{title}{Reconstruction of human movement trajectories from large-scale low-frequency mobile phone data},
\newblock \bibinfo{journal}{Computers, Environment and Urban Systems} \bibinfo{volume}{77} (\bibinfo{year}{2019}) \bibinfo{pages}{101346}.
\bibitem[{Zheng et~al.(2010)Zheng, Zheng, Xie, and Yang}]{zheng2010collaborative}
\bibinfo{author}{V.~W. Zheng}, \bibinfo{author}{Y.~Zheng}, \bibinfo{author}{X.~Xie}, \bibinfo{author}{Q.~Yang},
\newblock \bibinfo{title}{Collaborative location and activity recommendations with gps history data},
\newblock in: \bibinfo{booktitle}{Proceedings of the 19th international conference on World wide web}, pp. \bibinfo{pages}{1029--1038}.
\bibitem[{Alexander et~al.(2015)Alexander, Jiang, Murga, and Gonz{\'a}lez}]{alexander2015origin}
\bibinfo{author}{L.~Alexander}, \bibinfo{author}{S.~Jiang}, \bibinfo{author}{M.~Murga}, \bibinfo{author}{M.~C. Gonz{\'a}lez},
\newblock \bibinfo{title}{Origin--destination trips by purpose and time of day inferred from mobile phone data},
\newblock \bibinfo{journal}{Transportation research part c: emerging technologies} \bibinfo{volume}{58} (\bibinfo{year}{2015}) \bibinfo{pages}{240--250}.
\bibitem[{Liu et~al.(2024)Liu, Liao, Ma, He, Stanford, and Ma}]{liu2024human}
\bibinfo{author}{Y.~Liu}, \bibinfo{author}{X.~Liao}, \bibinfo{author}{H.~Ma}, \bibinfo{author}{B.~Y. He}, \bibinfo{author}{C.~Stanford}, \bibinfo{author}{J.~Ma},
\newblock \bibinfo{title}{Human mobility modeling with limited information via large language models},
\newblock \bibinfo{journal}{arXiv preprint arXiv:2409.17495}  (\bibinfo{year}{2024}).
\bibitem[{Cai et~al.(2024)Cai, Liu, Zhou, Ma, Zhao, Wu, and Ma}]{cai2024driving}
\bibinfo{author}{T.~Cai}, \bibinfo{author}{Y.~Liu}, \bibinfo{author}{Z.~Zhou}, \bibinfo{author}{H.~Ma}, \bibinfo{author}{S.~Z. Zhao}, \bibinfo{author}{Z.~Wu}, \bibinfo{author}{J.~Ma},
\newblock \bibinfo{title}{Driving with regulation: Interpretable decision-making for autonomous vehicles with retrieval-augmented reasoning via llm},
\newblock \bibinfo{journal}{arXiv preprint arXiv:2410.04759}  (\bibinfo{year}{2024}).
\bibitem[{Touvron et~al.(2023)Touvron, Lavril, Izacard, Martinet, Lachaux, Lacroix, Rozi{\`e}re, Goyal, Hambro, Azhar et~al.}]{touvron2023llama}
\bibinfo{author}{H.~Touvron}, \bibinfo{author}{T.~Lavril}, \bibinfo{author}{G.~Izacard}, \bibinfo{author}{X.~Martinet}, \bibinfo{author}{M.-A. Lachaux}, \bibinfo{author}{T.~Lacroix}, \bibinfo{author}{B.~Rozi{\`e}re}, \bibinfo{author}{N.~Goyal}, \bibinfo{author}{E.~Hambro}, \bibinfo{author}{F.~Azhar}, et~al.,
\newblock \bibinfo{title}{Llama: Open and efficient foundation language models},
\newblock \bibinfo{journal}{arXiv preprint arXiv:2302.13971}  (\bibinfo{year}{2023}).
\bibitem[{Anil et~al.(2023)Anil, Dai, Firat, Johnson, Lepikhin, Passos, Shakeri, Taropa, Bailey, Chen et~al.}]{anil2023palm}
\bibinfo{author}{R.~Anil}, \bibinfo{author}{A.~M. Dai}, \bibinfo{author}{O.~Firat}, \bibinfo{author}{M.~Johnson}, \bibinfo{author}{D.~Lepikhin}, \bibinfo{author}{A.~Passos}, \bibinfo{author}{S.~Shakeri}, \bibinfo{author}{E.~Taropa}, \bibinfo{author}{P.~Bailey}, \bibinfo{author}{Z.~Chen}, et~al.,
\newblock \bibinfo{title}{Palm 2 technical report},
\newblock \bibinfo{journal}{arXiv preprint arXiv:2305.10403}  (\bibinfo{year}{2023}).
\bibitem[{Li et~al.(2024)Li, Gou, Zhang, Liu, Li, Hu, Ling, Zhang, and Zhao}]{li2024teg}
\bibinfo{author}{Z.~Li}, \bibinfo{author}{Z.~Gou}, \bibinfo{author}{X.~Zhang}, \bibinfo{author}{Z.~Liu}, \bibinfo{author}{S.~Li}, \bibinfo{author}{Y.~Hu}, \bibinfo{author}{C.~Ling}, \bibinfo{author}{Z.~Zhang}, \bibinfo{author}{L.~Zhao},
\newblock \bibinfo{title}{Teg-db: A comprehensive dataset and benchmark of textual-edge graphs},
\newblock \bibinfo{journal}{arXiv preprint arXiv:2406.10310}  (\bibinfo{year}{2024}).
\bibitem[{Howard and Ruder(2018)}]{howard2018universal}
\bibinfo{author}{J.~Howard}, \bibinfo{author}{S.~Ruder},
\newblock \bibinfo{title}{Universal language model fine-tuning for text classification},
\newblock \bibinfo{journal}{arXiv preprint arXiv:1801.06146}  (\bibinfo{year}{2018}).
\bibitem[{Peters et~al.(2019)Peters, Ruder, and Smith}]{peters2019tune}
\bibinfo{author}{M.~E. Peters}, \bibinfo{author}{S.~Ruder}, \bibinfo{author}{N.~A. Smith},
\newblock \bibinfo{title}{To tune or not to tune? adapting pretrained representations to diverse tasks},
\newblock \bibinfo{journal}{arXiv preprint arXiv:1903.05987}  (\bibinfo{year}{2019}).
\bibitem[{Merchant et~al.(2020)Merchant, Rahimtoroghi, Pavlick, and Tenney}]{merchant2020happens}
\bibinfo{author}{A.~Merchant}, \bibinfo{author}{E.~Rahimtoroghi}, \bibinfo{author}{E.~Pavlick}, \bibinfo{author}{I.~Tenney},
\newblock \bibinfo{title}{What happens to bert embeddings during fine-tuning?},
\newblock \bibinfo{journal}{arXiv preprint arXiv:2004.14448}  (\bibinfo{year}{2020}).
\bibitem[{Liu et~al.(2019)Liu, Ott, Goyal, Du, Joshi, Chen, Levy, Lewis, Zettlemoyer, and Stoyanov}]{liu2019roberta}
\bibinfo{author}{Y.~Liu}, \bibinfo{author}{M.~Ott}, \bibinfo{author}{N.~Goyal}, \bibinfo{author}{J.~Du}, \bibinfo{author}{M.~Joshi}, \bibinfo{author}{D.~Chen}, \bibinfo{author}{O.~Levy}, \bibinfo{author}{M.~Lewis}, \bibinfo{author}{L.~Zettlemoyer}, \bibinfo{author}{V.~Stoyanov},
\newblock \bibinfo{title}{Roberta: A robustly optimized bert pretraining approach},
\newblock \bibinfo{journal}{arXiv preprint arXiv:1907.11692}  (\bibinfo{year}{2019}).
\bibitem[{Liu et~al.(2024)Liu, Kuai, Ma, Liao, He, and Ma}]{liuSemantic2024}
\bibinfo{author}{Y.~Liu}, \bibinfo{author}{C.~Kuai}, \bibinfo{author}{H.~Ma}, \bibinfo{author}{X.~Liao}, \bibinfo{author}{B.~Y. He}, \bibinfo{author}{J.~Ma},
\newblock \bibinfo{title}{Semantic trajectory data mining with llm-informed poi classification},
\newblock \bibinfo{journal}{arXiv preprint arXiv:2405.11715}  (\bibinfo{year}{2024}).
\bibitem[{Schneider et~al.(2013)Schneider, Belik, Couronn{\'e}, Smoreda, and Gonz{\'a}lez}]{schneider2013unravelling}
\bibinfo{author}{C.~M. Schneider}, \bibinfo{author}{V.~Belik}, \bibinfo{author}{T.~Couronn{\'e}}, \bibinfo{author}{Z.~Smoreda}, \bibinfo{author}{M.~C. Gonz{\'a}lez},
\newblock \bibinfo{title}{Unravelling daily human mobility motifs},
\newblock \bibinfo{journal}{Journal of The Royal Society Interface} \bibinfo{volume}{10} (\bibinfo{year}{2013}) \bibinfo{pages}{20130246}.
\bibitem[{Cao et~al.(2019)Cao, Li, Tu, and Wang}]{cao2019characterizing}
\bibinfo{author}{J.~Cao}, \bibinfo{author}{Q.~Li}, \bibinfo{author}{W.~Tu}, \bibinfo{author}{F.~Wang},
\newblock \bibinfo{title}{Characterizing preferred motif choices and distance impacts},
\newblock \bibinfo{journal}{Plos one} \bibinfo{volume}{14} (\bibinfo{year}{2019}) \bibinfo{pages}{e0215242}.
\bibitem[{Vaswani et~al.(2017)Vaswani, Shazeer, Parmar, Uszkoreit, Jones, Gomez, Kaiser, and Polosukhin}]{vaswani2017attention}
\bibinfo{author}{A.~Vaswani}, \bibinfo{author}{N.~Shazeer}, \bibinfo{author}{N.~Parmar}, \bibinfo{author}{J.~Uszkoreit}, \bibinfo{author}{L.~Jones}, \bibinfo{author}{A.~N. Gomez}, \bibinfo{author}{{\L}.~Kaiser}, \bibinfo{author}{I.~Polosukhin},
\newblock \bibinfo{title}{Attention is all you need},
\newblock \bibinfo{journal}{Advances in neural information processing systems} \bibinfo{volume}{30} (\bibinfo{year}{2017}).
\bibitem[{M{\"u}ller(2007)}]{muller2007dynamic}
\bibinfo{author}{M.~M{\"u}ller},
\newblock \bibinfo{title}{Dynamic time warping},
\newblock \bibinfo{journal}{Information retrieval for music and motion}  (\bibinfo{year}{2007}) \bibinfo{pages}{69--84}.
\bibitem[{Yosinski et~al.(2014)Yosinski, Clune, Bengio, and Lipson}]{yosinski2014transferable}
\bibinfo{author}{J.~Yosinski}, \bibinfo{author}{J.~Clune}, \bibinfo{author}{Y.~Bengio}, \bibinfo{author}{H.~Lipson},
\newblock \bibinfo{title}{How transferable are features in deep neural networks?},
\newblock \bibinfo{journal}{Advances in neural information processing systems} \bibinfo{volume}{27} (\bibinfo{year}{2014}).
\bibitem[{{Uber Engineering}(2018)}]{uber_h3_2018}
\bibinfo{author}{{Uber Engineering}}, \bibinfo{title}{{H3}: Uber’s hexagonal hierarchical spatial index}, \bibinfo{howpublished}{\url{https://www.uber.com/blog/h3/}}, \bibinfo{year}{2018}. \bibinfo{note}{Accessed 5 July 2025}.
\bibitem[{{US Bureau of Labor Statistics}(2023)}]{LAEmployment2023}
\bibinfo{author}{{US Bureau of Labor Statistics}}, \bibinfo{title}{County employment and wages in california — fourth quarter 2023}, \bibinfo{year}{2023}. \bibinfo{note}{Accessed: 2025-01-26}.
\bibitem[{Liao et~al.(2024)Liao, Liu, Kuai, Ma, He, Cao, Stanford, and Ma}]{liao2024reconstructing}
\bibinfo{author}{X.~Liao}, \bibinfo{author}{Y.~Liu}, \bibinfo{author}{C.~Kuai}, \bibinfo{author}{H.~Ma}, \bibinfo{author}{Y.~He}, \bibinfo{author}{S.~Cao}, \bibinfo{author}{C.~Stanford}, \bibinfo{author}{J.~Ma},
\newblock \bibinfo{title}{Reconstructing human mobility pattern: A semi-supervised approach for cross-dataset transfer learning},
\newblock \bibinfo{journal}{arXiv preprint arXiv:2410.03788}  (\bibinfo{year}{2024}).
\bibitem[{Luca et~al.(2021)Luca, Barlacchi, Lepri, and Pappalardo}]{luca2021survey}
\bibinfo{author}{M.~Luca}, \bibinfo{author}{G.~Barlacchi}, \bibinfo{author}{B.~Lepri}, \bibinfo{author}{L.~Pappalardo},
\newblock \bibinfo{title}{A survey on deep learning for human mobility},
\newblock \bibinfo{journal}{ACM Computing Surveys (CSUR)} \bibinfo{volume}{55} (\bibinfo{year}{2021}) \bibinfo{pages}{1--44}.
\bibitem[{Shumailov et~al.(2024)Shumailov, Shumaylov, Zhao, Papernot, Anderson, and Gal}]{shumailov2024ai}
\bibinfo{author}{I.~Shumailov}, \bibinfo{author}{Z.~Shumaylov}, \bibinfo{author}{Y.~Zhao}, \bibinfo{author}{N.~Papernot}, \bibinfo{author}{R.~Anderson}, \bibinfo{author}{Y.~Gal},
\newblock \bibinfo{title}{Ai models collapse when trained on recursively generated data},
\newblock \bibinfo{journal}{Nature} \bibinfo{volume}{631} (\bibinfo{year}{2024}) \bibinfo{pages}{755--759}.
\bibitem[{Bricka et~al.(2012)Bricka, Paleti, and Bhat}]{BrikaAnalysis2012}
\bibinfo{author}{S.~Bricka}, \bibinfo{author}{R.~Paleti}, \bibinfo{author}{C.~R. Bhat},
\newblock \bibinfo{title}{An analysis of the factors influencing differences in survey-reported and gps-recorded trips},
\newblock \bibinfo{journal}{Transportation Research Part C: Emerging Technologies} \bibinfo{volume}{21} (\bibinfo{year}{2012}) \bibinfo{pages}{67--78}.
\bibitem[{Wang et~al.(2024)Wang, Fang, Zeng, and Cheng}]{wang2024llmmob}
\bibinfo{author}{X.~Wang}, \bibinfo{author}{M.~Fang}, \bibinfo{author}{Z.~Zeng}, \bibinfo{author}{T.~Cheng},
\newblock \bibinfo{title}{Where would i go next? large language models as human mobility predictors},
\newblock \bibinfo{journal}{arXiv preprint arXiv:2308.15197}  (\bibinfo{year}{2024}).
\bibitem[{Feng et~al.(2020)Feng, Yang, Xu, Yu, Wang, and Li}]{feng2020movesim}
\bibinfo{author}{J.~Feng}, \bibinfo{author}{Z.~Yang}, \bibinfo{author}{F.~Xu}, \bibinfo{author}{H.~Yu}, \bibinfo{author}{M.~Wang}, \bibinfo{author}{Y.~Li},
\newblock \bibinfo{title}{Learning to simulate human mobility},
\newblock in: \bibinfo{booktitle}{Proc. 26th ACM SIGKDD Conf. on Knowledge Discovery and Data Mining}, pp. \bibinfo{pages}{3426--3434}.
\bibitem[{Zhu et~al.(2023)Zhu, Ye, Zhang, Zhao, and Yu}]{zhu2023difftraj}
\bibinfo{author}{Y.~Zhu}, \bibinfo{author}{Y.~Ye}, \bibinfo{author}{S.~Zhang}, \bibinfo{author}{X.~Zhao}, \bibinfo{author}{J.~J.~Q. Yu},
\newblock \bibinfo{title}{Difftraj: Generating gps trajectory with diffusion probabilistic model},
\newblock \bibinfo{journal}{arXiv preprint arXiv:2304.11582}  (\bibinfo{year}{2023}).
\bibitem[{Xia et~al.(2021)Xia, Qi, Feng, Xu, Sun, Guo, and Li}]{xia2021attnmove}
\bibinfo{author}{T.~Xia}, \bibinfo{author}{Y.~Qi}, \bibinfo{author}{J.~Feng}, \bibinfo{author}{F.~Xu}, \bibinfo{author}{F.~Sun}, \bibinfo{author}{D.~Guo}, \bibinfo{author}{Y.~Li},
\newblock \bibinfo{title}{Attnmove: History enhanced trajectory recovery via attentional network},
\newblock in: \bibinfo{booktitle}{Proc. AAAI Conf. on Artificial Intelligence}, volume~\bibinfo{volume}{35}, pp. \bibinfo{pages}{447--455}.
\bibitem[{Ren et~al.(2021)Ren, Ruan, Li, Bao, Meng, Li, and Zheng}]{ren2021mtrajrec}
\bibinfo{author}{H.~Ren}, \bibinfo{author}{S.~Ruan}, \bibinfo{author}{Y.~Li}, \bibinfo{author}{J.~Bao}, \bibinfo{author}{C.~Meng}, \bibinfo{author}{R.~Li}, \bibinfo{author}{Y.~Zheng},
\newblock \bibinfo{title}{Mtrajrec: Map-constrained trajectory recovery via seq2seq multi-task learning},
\newblock in: \bibinfo{booktitle}{Proc. 27th ACM SIGKDD Conf. on Knowledge Discovery and Data Mining}, pp. \bibinfo{pages}{1410--1419}.
\bibitem[{Si et~al.(2023)Si, Yang, Xiang, Wang, Li, Zhang, Tu, and Chen}]{si2023trajbert}
\bibinfo{author}{J.~Si}, \bibinfo{author}{J.~Yang}, \bibinfo{author}{Y.~Xiang}, \bibinfo{author}{H.~Wang}, \bibinfo{author}{L.~Li}, \bibinfo{author}{R.~Zhang}, \bibinfo{author}{B.~Tu}, \bibinfo{author}{X.~Chen},
\newblock \bibinfo{title}{Trajbert: Bert-based trajectory recovery with spatial-temporal refinement for implicit sparse trajectories},
\newblock \bibinfo{journal}{IEEE Transactions on Mobile Computing}  (\bibinfo{year}{2023}) \bibinfo{pages}{1--12}.
\bibitem[{Stopher et~al.(2010)Stopher, Xu, and Fitzgerald}]{StopherSydney2010}
\bibinfo{author}{P.~Stopher}, \bibinfo{author}{M.~Xu}, \bibinfo{author}{C.~Fitzgerald},
\newblock \bibinfo{title}{Assessing the accuracy of the sydney household travel survey with gps},
\newblock \bibinfo{journal}{Australasian Transport Research Forum} \bibinfo{volume}{33} (\bibinfo{year}{2010}) \bibinfo{pages}{1--18}.
\bibitem[{Gruber and Stark(2024)}]{GruberUnderreport2024}
\bibinfo{author}{C.~Gruber}, \bibinfo{author}{J.~Stark},
\newblock \bibinfo{title}{Underreported trips, a non-negligible empirical effect of traditional survey methods},
\newblock \bibinfo{journal}{Transportation Research Procedia} \bibinfo{volume}{69} (\bibinfo{year}{2024}) \bibinfo{pages}{312--320}.

\end{thebibliography}


\end{document}